%% file: main.tex
\documentclass[journal]{IEEEtran}
\usepackage{hyperref}
\usepackage{cleveref}

\usepackage{cite}

\ifCLASSINFOpdf
\else
\fi
\usepackage{graphicx}
\usepackage{tabu}
\usepackage{array}
\usepackage{multirow}
\usepackage{float}
\usepackage{subfig}
\usepackage{amsmath}
\usepackage{color}
\usepackage[table]{xcolor}
\usepackage{fancyhdr}

\newcommand{\ssmall}{\fontsize{6pt}{\baselineskip}\selectfont}

\begin{document}
%
\title{Saliency Integration: An Arbitrator Model}
%
%
%

\author{Yingyue~Xu*,
        Xiaopeng~Hong*,
        Fatih Porikli,~\IEEEmembership{Fellow,~IEEE},
        Xin~Liu,
        Jie~Chen,   and~Guoying~Zhao\dag,~\IEEEmembership{Senior~Member,~IEEE}

\IEEEcompsocitemizethanks{\IEEEcompsocthanksitem * The first two authors contributed equally. \dag Corresponding author.}
\IEEEcompsocitemizethanks{\IEEEcompsocthanksitem
Yingyue Xu, Xiaopeng Hong, Xin Liu, Jie Chen, and Guoying Zhao are with the Center for Machine Vision and Signal Analysis, University of Oulu, Finland. Emails: \{yingyue.xu, xiaopeng.hong, xin.liu, jie.chen, guoying.zhao\}@oulu.fi.
Fatih Porikli is with the Research School of Engineering, Australian National University, Canberra. Email: fatih.porikli@anu.edu.au.}}

\maketitle

\thispagestyle{fancy} 
\fancyhead{} 
\lhead{Accepted to IEEE Transactions on Multimedia 2018} 
\renewcommand{\headrulewidth}{0pt}

\pagestyle{fancy}
\lhead{Accepted to IEEE Transactions on Multimedia 2018} 
\renewcommand{\headrulewidth}{0pt}

\begin{abstract}
Saliency integration has attracted much attention on unifying saliency maps from multiple saliency models. Previous offline integration methods usually face two challenges:
1. if most of the candidate saliency models misjudge the saliency on an image, the integration result will lean heavily on those inferior candidate models; 2. an unawareness of the ground truth saliency labels brings difficulty in estimating the expertise of each candidate model. To address these problems, in this paper, we propose an arbitrator model (AM) for saliency integration.
Firstly, we incorporate the consensus of multiple saliency models and the external knowledge into a reference map to effectively rectify the misleading by candidate models.
Secondly, our quest for ways of estimating the expertise of the saliency models without ground truth labels gives rise to two distinct online model-expertise estimation methods.
Finally, we derive a Bayesian integration framework to reconcile the saliency models of varying expertise and the reference map.
To extensively evaluate the proposed AM model, we test twenty-seven state-of-the-art saliency models, covering both traditional and deep learning ones, on various combinations over four datasets. The evaluation results show that the AM model improves the performance substantially compared to the existing state-of-the-art integration methods, regardless of the chosen candidate saliency models.

\end{abstract}

\begin{IEEEkeywords}
saliency integration, saliency aggregation, online model, arbitrator model.
\end{IEEEkeywords}

%
\IEEEpeerreviewmaketitle

\input{Introduction.tex}

\section{Arbitrator Model}
\label{sec:model}
In this paper, we propose a Bayesian framework, namely the Arbitrator Model (AM), for saliency integration. As illustrated in Fig.~\ref{fig:framework}, the AM model takes the test image and the corresponding saliency heat maps obtained by $P$ saliency models as input. It consists of two main units: 1) a reference generator which makes use of the consensus of the input heat maps and the external knowledge; 2) an online estimator which treats the $P$ saliency maps as candidates and evaluates their corresponding qualities (as termed by \emph{expertise} hereinafter).

In the following of this section, we will derive the framework of the Arbitrator model (Section~\ref{subsec:framework}) and provide an efficient solution for integration (Section ~\ref{subsec:CA}). The reference generator and the online estimator will be detailed in Sections~\ref{sec:prior}) and \ref{sec:expertise}) respectively.

\input{framework.tex}

\input{CA.tex}

\input{Prior.tex}

\input{Expertise.tex}

\input{experiments.tex}

\input{Conclusion.tex}


%

\section*{Acknowledgment}
The authors are grateful to the Academy of Finland, Tekes Fidipro Program (Grant No. 1849/31/2015), Business Finland project (Grant No.3116/31/2017), Infotech, and Natural Science Foundation of China  (Grant No. 61772419/61572205/61601362). The authors also wish to acknowledge CSC-IT Center for Science, Finland, for generous computational resources,
and the supports of NVIDIA Corporation with the donation of the GPUs used for this research. Besides, we appreciate the comments and codes from Prof. Huchuan Lu and Mr. Mengyang Feng.

\ifCLASSOPTIONcaptionsoff
  \newpage
\fi



%
\bibliographystyle{IEEEtran}
\bibliography{egbib}

%





\end{document}

%% file: Introduction.tex
\section{Introduction}\label{sec:introduction}
\IEEEPARstart{O}{ver} the past two decades, saliency detection has hit much attention for its broad applications, such as image and video segmentation~\cite{rahtu2010segmenting}, video compression~\cite{guo2010novel}, and advertising~\cite{hua2008multimedia}.
Aiming at highlighting the regions of interest (ROI) of the human visual system on a scene with biologically plausible cues, a variety of saliency models have been proposed~\cite{itti1998model,harel2006graph,walther2006modeling,bruce2005saliency,hou2007saliency,achanta2009frequency,judd2009learning,jiang2014saliency,le2006coherent,zhang2008sun,cheng2015global,lu2014learning,zhu2014saliency,shi2014reverse,liu2014adaptive,qin2015saliency,liu2015predicting,zhao2015saliency,tong2015salient,gong2015saliency,zhang2015co,feng2016fixation,xu2015task}.

Existing saliency models utilize a broad range of strategies such as coarse-to-fine saliency map estimation~\cite{gong2015saliency,wang2015deep}, top-down~\cite{judd2009learning,kanan2009sun} or bottom-up~\cite{itti1998model,harel2006graph,walther2006modeling,bruce2005saliency,hou2007saliency,erdem2013visual,margolin2013makes,qin2015saliency} feature extraction, making different assumptions, for example, the background surrounding assumption~\cite{hou2012image,liu2014adaptive,zhu2014saliency,shen2012unified,qin2015saliency,xu2018saliency}, and relying on a variety of models including support vector machine~\cite{judd2009learning}, AdaBoost~\cite{tong2015salient}, multiple kernel learning~\cite{jiang2014saliency,shen2014webpage}, and deep convolutional neural networks~\cite{liu2015predicting,zhao2015saliency,zhang2015co,bruce2016deeper,pan2016shallow}, \emph{etc}.

\begin{figure*}[t]
\begin{center}
\scalebox{1}{\includegraphics[width=7in]{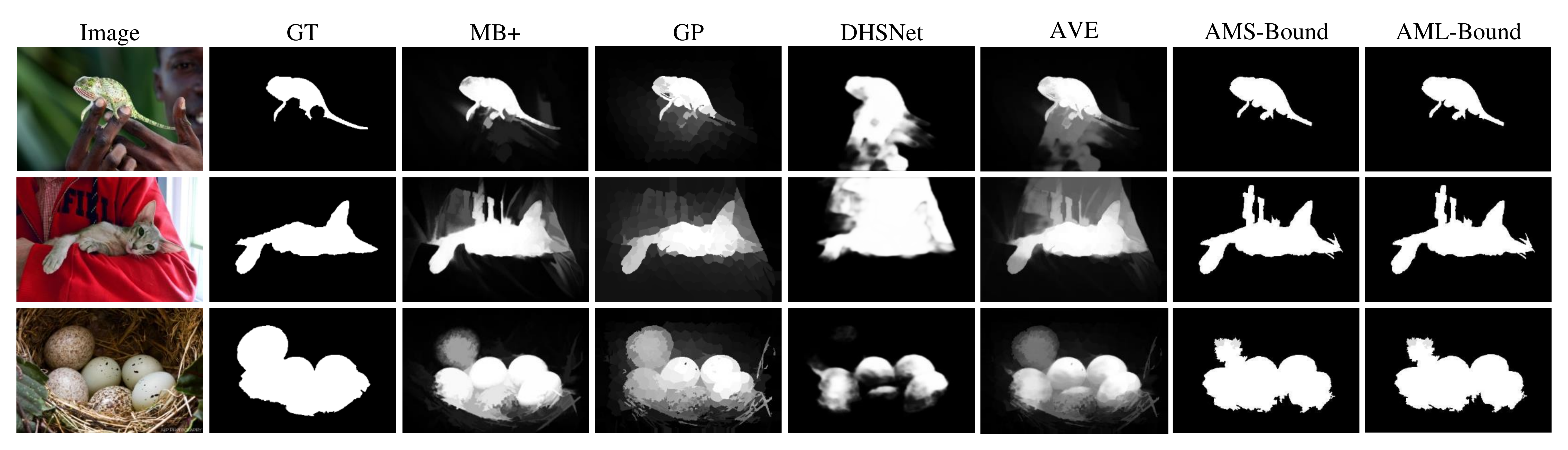}}
\caption{Examples where saliency maps from the state-of-the-art deep model show inferior predictions to the traditional models. From left to right there are original images, ground truth (GT), traditional saliency models \textit{e.g.}, MB+~\cite{zhang2015minimum} and GP~\cite{jiang2015generic}, deep saliency model \textit{e.g.}, DHSNet~\cite{liu2016dhsnet}, naive integration approach \textit{e.g.}, average map (AVE), and our proposed arbitrator model (AMS-Bound and AML-Bound). Examples are selected from the ECSSD dataset.}
\label{fig:BestSingleVsInte}
\end{center}
\end{figure*}

Recently, saliency integration (or saliency aggregation) approaches, which unify saliency maps from multiple existing saliency models, have attracted much attention~\cite{le2014saliency,song2012saliency,mai2013saliency,wang2016learning,borji2012salient,qin2015saliency}. Although many of modern saliency models claim high performance in the statistical sense on different public benchmarks, none of them can outperform the others for every image under evaluation~\cite{mai2014comparing,borji2015salient}.
For instance, even though the deep DHSNet~\cite{liu2016dhsnet} model, as one of the state-of-the-art approaches, is usually considered to surpass the traditional methods \textit{e.g.}, GP~\cite{jiang2015generic}, and MB+~\cite{zhang2015minimum}, there are still images where DHSNet shows inferior predictions to GP and MB+, as shown in Figure~\ref{fig:BestSingleVsInte}. Thus, saliency (heat map) integration is proposed to take the advantages of multiple saliency models and make up for the defects of any specific ones, for enhanced accuracy and robustness of saliency detection.

Saliency integration is essentially a weighted combination of multiple saliency maps~\cite{le2014saliency}. The weights, assuredly, play a central role in saliency integration.
According to different ways of setting the weights, existing saliency integration approaches can be briefly categorized into the following two types.

\emph{Offline} saliency integration models weigh candidate models by optimizing a specific energy function using a collection of data prepared in advance~\cite{le2014saliency,song2012saliency,mai2013saliency,wang2016learning}. The integration model is fixed once the learning phase has been completed. However, they usually require extra efforts in providing the training samples with ground truth labels.
Moreover, the scalability is limited~\cite{remi2016svm}, as the parameter settings of the integration model are only valid for a particular combination of candidate models. If a new candidate model is added, the integration model has to be retrained.
Furthermore, there is an underline assumption that the \emph{known} samples for learning and the \emph{unknown} samples for prediction possess similar distributions. If the distribution of the \emph{unknown} samples is significantly different from those of the \emph{known} ones, the learnt parameters may fail in prediction.

\emph{Online} integration models~\cite{le2014saliency,borji2012salient,qin2015saliency} are brought forth as a means of addressing the aforementioned problems of the \emph{offline} models.
\emph{Online} models determine the weights of saliency maps by adapting to the image under evaluation directly, without the demand of any (pre-)collection of \emph{known} samples. The resulting weights are, therefore, image-specific. Compared to the \emph{offline} models, the \emph{online} ones are free from fixing a model in advance and thus much more flexible and efficient. However, as every coin has two sides, online models have to face \textbf{two main challenges}.


1. \textbf{How to efficiently estimate the expertise of candidate models?}
Most of the previous works assume that the expertise (\emph{a.k.a.}, weights or contribution) of each candidate saliency model is equal (\emph{e.g.}, BN~\cite{borji2012salient} and MCA~\cite{qin2015saliency}).
This assumption greatly eases the computational burden. However, it loses sight of the fact that each candidate saliency model shows discrepant ability in predicting an image.
In fact, the performance of an integration without consideration of the expertise of candidate models may decrease,
as the voices of the superior models are easily drowned out by the mistakes made by those inferior ones.
However, it is extremely difficult for online models to weigh each saliency model accurately, since there is no supervised information of the test images.
Mai \emph{et al.}~\cite{mai2014comparing} proposed to rank the performances of the saliency models on an image without the ground truth. However, since the ranking is a sequence of ordinal numbers, it cannot numerically measure the performance of each saliency model on the image in details. 
Le Meur \emph{et al.}~\cite{le2014saliency} 
estimate the expertise of the candidate saliency models 
by a weight function called M-estimator. 
The M-estimator decreases the expertise of the outliers that are detected according to their distances to a linear summation
of the candidate saliency maps.
However, as shown by the experimental results~\cite{le2014saliency},
the M-estimators perform similarly to \emph{average weighting}, indicating that the computed weights are far from accurately specifying the expertise of the candidate models. Recently, some integration approaches~\cite{zhang2017supervision,Quan2017Unsupervised} explore expertise estimation by bringing the concept of superpixel difficulty, as each superpixel of an image may possess different difficulty for saliency assessment. This concept of using superpixel difficulty together with model expertise as hidden variables facilitates the expertise estimation process from a more refined superpixel level.

\begin{figure*}[t]
\begin{center}
\scalebox{1}{\includegraphics[width=7in]{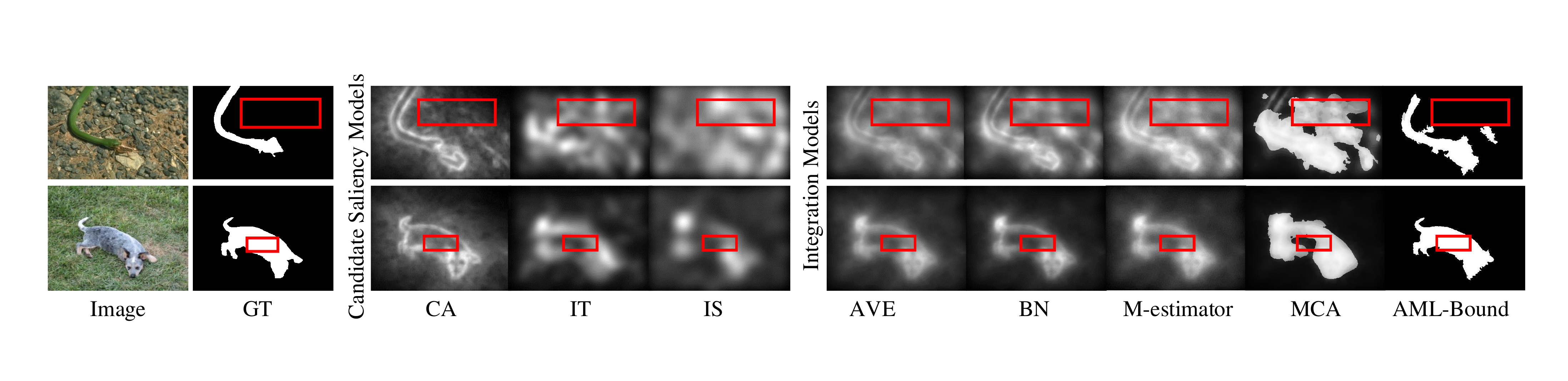}}
\vspace*{-6mm}
\caption{Examples of misleading caused by misjudgement from candidate saliency models. From left to right columns are original images, ground truth (GT), candidate saliency models including CA~\cite{goferman2012context}, IT~\cite{itti1998model}, IS~\cite{hou2012image}, average map (AVE), integrated maps of BN~\cite{borji2012salient}, M-estimator~\cite{le2014saliency}, MCA~\cite{qin2015saliency}, and our proposed arbitrator model (AML-Bound). The red rectangles on GT indicate the misjudged regions by the candidate saliency models.}
\label{fig:mismeasure}
\end{center}
\end{figure*}

2. \textbf{How to ensure solid performance enhancement?} Le Meur \emph{et al.}~\cite{le2014saliency} also indicate that saliency integration models may decrease the performance in many cases. For instance, when most candidate saliency models misjudge a region on an image, the integration result will be highly susceptible to error. In Figure~\ref{fig:mismeasure}, we present the integration maps given by four typical online integration models using three popular saliency candidate methods on two images. The red rectangles on the ground truth indicate the regions that the candidate saliency models misjudge. From the integrated saliency maps, it can be perceived that when candidate saliency models misjudge a region on an image, the region will also be misjudged on the integrated map. Thus, overcoming the misleading by most of the candidate saliency models for solid performance enhancement becomes another big challenge in saliency integration.

This paper focuses on \emph{online} saliency integration methods to treat the two challenges simultaneously:

(1) The saliency integration approach should efficiently determine the expertise of each candidate saliency model, in an online manner.

(2) The saliency integration method should have a mechanism to rectify the misleading by candidate models, even if most of the models misjudge a region on an image.

Based on the above two principles, we propose an online saliency integration framework, which is termed by the arbitrator model (AM) in this paper, as illustrated in Figure~\ref{fig:framework}. We derive a Bayesian framework with the following two main components to reconcile the principles:

(1) We incorporate the consensus of multiple saliency models and the external knowledge into a reference map to effectively rectify the misleading by candidate models.

(2) Our quest for ways of estimating the expertise of the saliency models without ground truth labels gives rise to two distinct online model-expertise estimation methods: One is a statistical approach and the other is latent-variable-based. The two methods measure the expertise of the candidate saliency models without supervised information of the given test image, and meet the requirements of computing rational expertise of the candidate models.

The contributions of this paper are three-fold:

(1) We propose a Bayesian saliency integration framework, which makes the most of clues from both candidate saliency models and the external knowledge.

(2) We explore online ways of measuring the expertise of multiple saliency models and successfully introduce two methods.

(3) To extensively evaluate the proposed AM model, we test twenty-seven state-of-the-art candidate saliency models, including both traditional and deep learning ones, on various combinations over four datasets. To our best knowledge, the number of candidates and combinations under evaluation are the largest (ones) among the state-of-the-art saliency integration works.

%% file: framework.tex
\subsection{Bayesian Integration Framework}
\label{subsec:framework}
Superpixel algorithms group the pixels on an image into perceptually consistent units, and thus reserve the essential local structure of the image. It also efficiently reduces the computational costs of the subsequent processing tasks. Thus, the AM model proceeds to unify multiple saliency maps in the level of superpixel instead of the pixel level.

Given an image of $N$ superpixels, each superpixel has a unique saliency label $ l_n \in \{0,1\}$. We define the events that the $n$-th superpixel is salient (foreground) and inconspicuous (background) by $F_n$ and $\bar{F}_n$ respectively. Apparently, we have $P \left ( {F_n} \right )=P\left ( l_n=1 \right )$, while $P\left ( \bar{F_n} \right )= 1-P \left ( {F_n} \right )=P\left ( l_n=0 \right )$.

Suppose there are $P$ 
{saliency models}, each model is able to assign a saliency intensity value $s_{p,n} \in \left [ 0,1 \right ]$ to the $n$-th superpixel on the $p$-th saliency map. The binary saliency label of the $n$-th superpixel by the $p$-th model, is denoted as $\iota_{p,n} \in \{0,1\}$. $\iota_{p,n} = 1$ indicates the $n$-th superpixel is considered as a foreground one by the $p$-th model and vice versa. It can be easily obtained via a binarization process on the saliency intensity $s_{p,n}$ with a threshold $\gamma_p$, \emph{e.g.}, OTSU thresholding~\cite{otsu1975threshold}. More specifically, we have $\iota_{p,n} = 1$ ($\boldsymbol{\iota}_{p,n}$), if $s_{p,n} \geq \gamma_p$, otherwise, $\iota_{p,n} = 0$. Similarly, $\iota_{{q\neq p},n} = 1$ ($\boldsymbol{\iota}_{{q\neq p},n}$), if $s_{{q\neq p},n} \geq \gamma_p$, otherwise, $\iota_{{q\neq p},n} = 0$. Given the intensity of the $n$-th superpixel from the $p$-th model $s_{p,n}$ and the $n$-th superpixel being labeled as foreground on the binary maps by the rest models $\boldsymbol{\iota}_{{q\neq p},n}$, the probability that the $n$-th superpixel is measured as foreground by the $p$-th model is $P\left ( F_n | s_{p,n}, \boldsymbol{\iota}_{{q\neq p},n} \right ).$\footnote[1]{In this work, although the contexts with respect to $F$, $\bar{F}$, $s$ and $\iota$ are upon a superpixel $n$, the sub-index $n$ is omitted for clarity unless otherwise specified.}

The probability $P\left ( F | s_p, \boldsymbol{\iota}_{q\neq p} \right )$ is derived under the Bayesian probability framework:
\begin{equation} \label{eq:postprob_F}
\begin{split}
P\left ( F | s_p, \boldsymbol{\iota}_{q\neq p} \right ) &\propto P\left ( F \right )P\left (  s_p, \boldsymbol{\iota}_{q\neq p} | F \right ) \\
&= P\left ( F \right )P\left ( s_p | F\right )P\left (  \boldsymbol{\iota}_{q\neq p} | s_p,F \right )\\
&=P\left ( F \right )P\left ( s_p | F \right ) \prod_{q\neq p}P\left ( \boldsymbol{\iota}_{q} | F \right ),
\end{split}
\end{equation}
\noindent with the assumption that all $P$ saliency models make decisions independently, either with respect to the saliency intensity $s$ or the binary saliency label $\iota$. $s_p$ represents the $p$-th saliency intensity map, while $\iota_p$ is the $p$-th binary saliency map.

The ratio of $\frac{P\left ( F | s_p, \boldsymbol{\iota}_{q\neq p} \right )}{P\left ( {\bar{F}} | s_p, \boldsymbol{\iota}_{q\neq p} \right )}$ is computed as follow:
\begin{equation} \label{eq:post_ratio}
\begin{split}
\Lambda \left ( F | s_p, \boldsymbol{\iota}_{q\neq p} \right )
&=\frac{P\left ( F | s_p, \boldsymbol{\iota}_{q\neq p} \right )}{P\left ( \bar{F} | s_p, \boldsymbol{\iota}_{q\neq p} \right )} \\
&=\frac{P\left ( F \right )}{P\left ( \bar{F} \right )}
\frac{P\left ( s_p | F \right ) }{P\left ( s_p | \bar{F}  \right ) }
\prod_{q\neq p}\frac{P\left (  \boldsymbol{\iota}_{q} | F \right )}{P\left (  \boldsymbol{\iota}_{q} | \bar{F}  \right )}
\end{split}
\end{equation}

Then we compute the logarithm function of $\Lambda\left ( F | s_p, \boldsymbol{\iota}_{q\neq p} \right )$ to form the integration framework, namely the arbitrator model (AM), as follow:
\begin{equation} \label{eq:ln_post_ratio}
\begin{split}
\ln&{\Lambda\left ( F | s_p, \boldsymbol{\iota}_{q\neq p} \right )}\\
=&\ln{\frac{P\left ( F \right )}{P\left ( \bar{F} \right )}}
+\ln{\frac{P\left ( s_p | F \right ) }{P\left ( s_p | \bar{F}  \right ) }}
+\sum_{q\neq p}\ln{\frac{P\left (  \boldsymbol{\iota}_{q} | F \right )}{P\left (  \boldsymbol{\iota}_{q} | \bar{F}  \right )}}\\
\end{split}
\end{equation}

\begin{figure*}[t]
\begin{center}
\scalebox{1}{\includegraphics[width=5.2in]{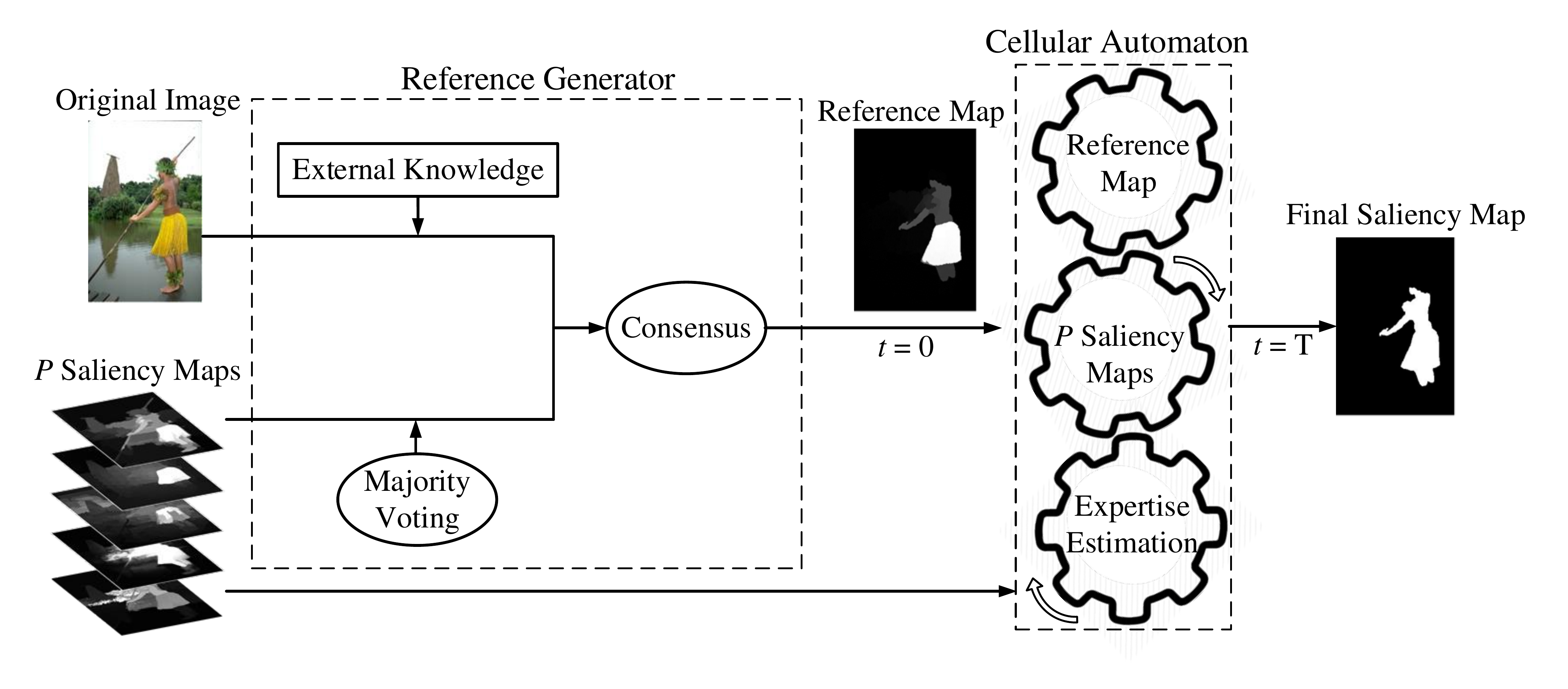}}
\vspace*{-5mm}
\caption{Framework of the proposed arbitrator model (AM). The arbitrator incorporates the consensus of multiple saliency maps and the external knowledge into a reference map via the reference generator. A Bayesian integration framework reconciles the reference map and \emph{P} saliency maps of varying expertise with cellular automaton (CA), to compute the final result. $\alpha_p$ and $\beta_p$ are the expertise of the $p$-th saliency intensity map and the $p$-th binary map respectively. After each generation of the CA, the \emph{P} saliency maps are updated. Accordingly, the expertise and the reference map are updated based on the new \emph{P} saliency maps.}
\label{fig:framework}
\end{center}
\end{figure*}

%% file: CA.tex
\subsection{Implementation}
\label{subsec:CA}



We incorporate all the terms in Eq.~\ref{eq:ln_post_ratio} into a cellular automaton to generate a final saliency map.

Cellular Automaton (CA), \emph{a.k.a.}, cellular space or homogeneous structure, is a discrete model in computability theory and mathematics~\cite{von1951general,wolfram1983statistical,qin2015saliency}. A CA consists of a regular grid of cells.
Each cell is with states, which are either discrete (\emph{e.g.}, `On' and `Off')~\cite{von1951general,wolfram1983statistical} or continuous (\emph{e.g.}, between 0 and 1)~\cite{qin2015saliency,qin2018saliency}. The neighborhood of one specific cell can influence the states of the specific cell in next generations (advancing \emph{t} by 1) in line with certain updating rules. Generally, the rule of updating the states of cells is a mathematical function, which is usually synchronous to all cells and time invariants.


Cellular automaton provides an efficient mechanism to propagate information to a batch of cells from their neighborhood respectively. For online saliency integration, we treat the fusion of the candidate saliency maps as a dynamic system by concentrating on the contextual relationships between different candidate saliency maps. More specifically, each unit (superpixels or pixels) of one candidate map is regarded as a cell and its corresponding saliency value as the `state'. The units with the same locations on the other candidate maps become the neighbors of the cell.
Considering the states within the neighborhood are contextually coherent, the states of the neighbors become indicators that orient one specific cell to evolve.
One specific cell thus intends to make a wise decision for its state in the next generation, depending on the current states of its own and its neighbors.
Intuitively, if its state is similar to the neighbors with high confidence, the cell should maintain its state; otherwise, the state should be updated towards the states of its high-confidence neighbors.
As a result, as the CA updates itself iteratively, the states (saliency values) of all cells evolve for better estimation of the saliency values to make the dynamic system more stable.

As $\Lambda\left ( F | s_p, \boldsymbol{\iota}_{q\neq p} \right ) = \frac{P\left ( F | s_p, \boldsymbol{\iota}_{q\neq p} \right )}{P\left ( {\bar{F}} | s_p, \boldsymbol{\iota}_{q\neq p} \right )} = \frac{P\left ( F | s_p, \boldsymbol{\iota}_{q\neq p} \right )}{1 - P\left ( F | s_p, \boldsymbol{\iota}_{q\neq p} \right ) )}$, the  left side of Eq.~\ref{eq:ln_post_ratio} is the logarithm of the posterior ratio $\Lambda\left ( F | s_p, \boldsymbol{\iota}_{q\neq p} \right )$ and thus a \emph{logit} function of $P\left ( F | s_p, \boldsymbol{\iota}_{q\neq p} \right )$. In this paper, $P\left ( F | s_p, \boldsymbol{\iota}_{q\neq p} \right )$ is defined as ${s}_p^{t+1}$, which stands for the saliency value (of the $n$-th superpixel) on the $p$-th saliency intensity map at time $t+1$.

There are three terms on the right side of Eq.~\ref{eq:ln_post_ratio}. 1) The first term $\ln{\frac{P\left ( F \right )}{P\left ( \bar{F} \right )}}=\ln{\frac{P\left ( F \right )}{1 - P\left ({F} \right )}}$ is a logit function of $P(F)$. It is defined as $\textup{logit}(S_{\textup{Ref}}^t)$, where $S_{\textup{Ref}}^t$ represents the saliency reference map at time $t$. The term at time $0$ is initialized as the reference map $S_{\textup{Ref}}^0$ which rectifies the misleading by candidate models. 2) The second term $\textup{ln}\frac{P\left ( s_p | F \right ) }{P\left ( s_p | \bar{F}  \right ) }$ is the logarithm of the ratio of marginal likelihoods $\frac{P\left ( s_p | F \right ) }{P\left ( s_p | \bar{F}  \right ) }$, which is clearly linked to the confidence or the reliability of the saliency intensity provided by the $p$-th candidate model. The second term is thus defined as ${\ln\left(\alpha_p^{t}\right)}\cdot s_p^{t}$, where $\alpha_p$ is the expertise of the $p$-th method and $s_p^{t}$ is the $p$-th saliency intensity map at time $t$. 3) Similarly, the third term $\textup{ln}\frac{P\left (  \boldsymbol{\iota}_{q} | F \right )}{P\left (  \boldsymbol{\iota}_{q} | \bar{F}  \right )}$ is associated with the confidence or the reliability of
the $q$-th binary saliency map. Thus, we adopt $\mathcal{E}(s_q^{t}-\gamma_q^t)$ to threshold the $q$-th saliency map to obtain the corresponding binary one. Denoting the expertise by $\beta_q$, we define
the third term as $\sum_{q\neq p}{\ln\left(\beta_q^{t}\right)\cdot \mathcal{E}(s_q^{t}-\gamma_q^t)}$.


At time $t=0$, the reference map $S_{\textup{Ref}}^0$ is initialized from a reference generator as in Section~\ref{sec:prior}. At each generation $t>0$, we possess the reference map $S_{\textup{Ref}}^t$ and the saliency maps ($s_p^t$) of varying expertise $\alpha_p^t$ and $\beta_p^t$. Then we adopt CA to compute the corresponding $S_{\textup{Ref}}^{t+1}$, $s_p^{t+1}$, $\alpha_p^{t+1}$ and $\beta_p^{t+1}$ at time $t+1$. The synchronous updating rule of the cellular automaton derived from Eq.~\ref{eq:ln_post_ratio} is as follows:
\begin{equation} \label{eq:updating_rule}
\begin{split}
\textup{logit}\left({s}_p^{t+1}\right )=
 &\textup{logit}\left(S_{\textup{Ref}}^t\right )\\
&+{\ln\left(\alpha_p^{t}\right)}\cdot s_p^{t}\\
&+\sum_{q\neq p}{\ln\left(\beta_q^{t}\right)\cdot \mathcal{E}(s_q^{t}-\gamma_q^t)}.
\end{split}
\end{equation}

\begin{equation} \label{eq:sprior}
S_{\textup{Ref}}^{t} = \frac{1}{P}\sum_{p=1}^{P}{s}_p^{t}.
\end{equation}


\begin{equation} \label{eq:final_saliency}
S_{\textup{Final}}=\frac{1}{P}\sum_{p=1}^{P} s_p^{T}.
\end{equation}

\begin{figure}
\begin{center}
\scalebox{1}{\includegraphics[width=3.2in]{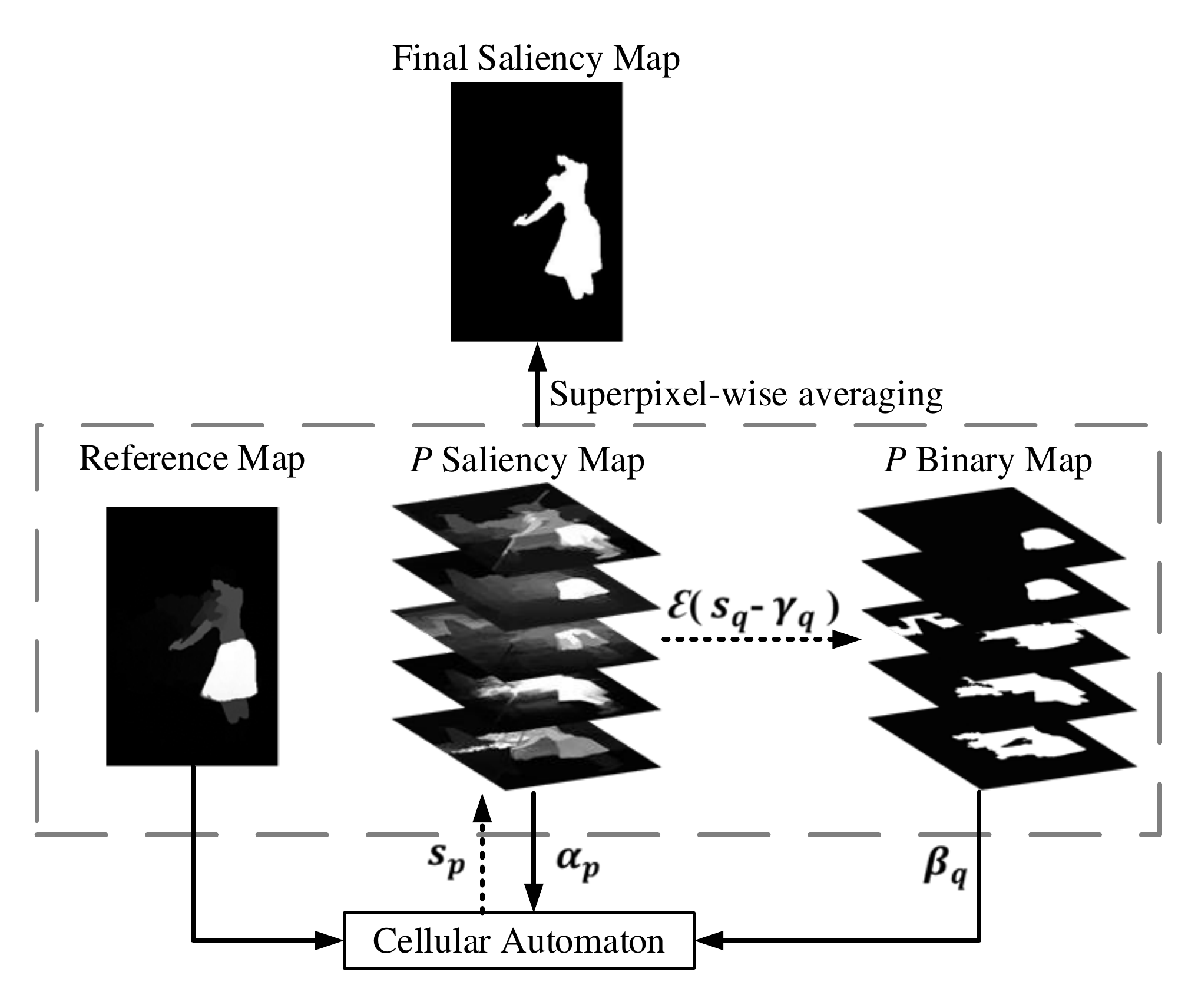}}
\caption{Cellular automaton in the proposed arbitrator model.
At each generation, the CA incorporates the reference map,
the $p$-th saliency intensity map $s_p$ with its expertise $\alpha_p$ and the rest binary maps with varying expertise $\beta_q$ ($q\neq p$) by the expertise estimator into the final saliency map, as shown by solid arrows. $s_p$ and the threshold $\gamma_p$ are updated at each generation as shown by dashed arrows. $\mathcal{E}(s_q-\gamma_q)$ thresholds the $q$-th saliency intensity map to get the binary map.}
\label{fig:CA}
\end{center}
\end{figure}

\noindent The updating process of cellular automaton is illustrated in Figure~\ref{fig:CA}. As empirically verified in experiments Section~\ref{sec:experiment}, all the saliency intensity maps ${s_p^{T}}, p = 1,\ldots, P$ will converge into stable states within 5 generations.

CA has been adopted for saliency integration in MCA~\cite{qin2015saliency}. However, there is a strong assumption that the expertise of candidate models should be equal, which limits the integration performance as discussed in Section~\ref{sec:introduction}.
In our arbitrator model, we largely enrich the work by incorporating the reference map and candidate models with varying expertise into CA, which is out of the range of the previous MCA.

We will further introduce the computation of the reference map $S_{\textup{Ref}}^0$, namely the reference generator, in Section~\ref{sec:prior}. Then we will detail the model-estimator that measures the expertise of saliency intensity maps $\alpha_p$ and the expertise of binary maps $\beta_q$ in Section~\ref{sec:expertise}.

%% file: Prior.tex
\section{Reference Generator}
\label{sec:prior}
As mentioned in Section~\ref{sec:introduction}, there is possible misleading by the candidate saliency models. To overcome this problem for solid performance enhancement, we propose to hear voices from both the candidate models and the external knowledge. The reference map $S_{\textup{Ref}}^0$, derived from $P(F)$ in Eq.~\ref{eq:ln_post_ratio}, provides a natural scheme to introduce the external knowledge about salient object detection.

To acquire the reference map, we firstly compute an external saliency intensity map with external knowledge. Then, we compute a consensus map based on the consistency of the candidate models and the external knowledge map. Finally, we propagate the consensus map to get the reference map.

\subsubsection{The External Knowledge Map}
\label{subsubsec:Ext}
We introduce the external knowledge map to rectify the errors by the candidate models. Basically, the external knowledge can be any reasonable assumptions about salient object detection or currently existing saliency models. However the selection of the external knowledge is critical to the final integration performance.

In this work, we investigate three distinct methods to compute the external knowledge map. The first one is a handy and fast method based on the widely accepted assumptions, such as boundary prior~\cite{wei2012geodesic,zhu2014saliency,qin2015saliency,li2013saliency,wang2016grab,jiang2013salient,wang2015deep,han2015background,zhang2017revealing}. The second is the saliency map from one of the state-of-the-art traditional saliency models such as CCM~\cite{liu2017hierarchical}. The third one is the saliency map from one of the state-of-the-art deep models such as DHSNet model~\cite{liu2016dhsnet}.

{$\bullet$} \textbf{Knowledge Map from Assumptions}

In recent saliency detection approaches, it is widely accepted that the boundaries of a given image are most likely to be the background regions. Wei \emph{et al.}~\cite{wei2012geodesic,zhu2014saliency} point out that the most background regions, other than salient ones, are easily connected to image boundaries. This boundary prior theory comes from basic rules of photographic composition, and even if salient objects are far from the center, they seldom touch image boundaries (validated on MSRA datasetset in ~\cite{wei2012geodesic}). Similarly, a number of saliency models~\cite{qin2015saliency,li2013saliency,wang2016grab} generated a coarse saliency map with the compactness of image boundaries. Besides, several supervised saliency models~\cite{jiang2013salient,wang2015deep,han2015background} also extracted the appearance features of boundaries for model training. Therefore, we compute the external map $S_{\textup{Ext}}$ based on this boundary prior knowledge.

We assume that the more discrepant a superpixel is from the boundary superpixels, the more salient the superpixel is. We describe the mean CIELab feature of the $n$-th superpixel on an image as $\{c_n\}$ and select the superpixels along the four boundaries as boundary seeds. The boundary seeds are grouped into $K$ clusters by K-means algorithm~\cite{hartigan1979algorithm}, $c_b^k$ is the boundary superpixel belonging to the $k$-th cluster, $K$ is empirically set as 3. For each superpixel $c_n$, we compute its appearance similarities to each cluster. If the superpixel is still quite different from its most similar cluster, it is more likely to be salient. In this way, we obtain the external knowledge map $S_{\textup{Ext}}$:

\begin{equation} \label{eq:backgroundprior}
S_{\textup{Ext}}(c_n) = \min_{k \in \{1,\ldots,K\}} ({\frac{1}{N_k}\sum_{b = 1}^{N_k} \lVert c_n-c_b^k \lVert}),
\end{equation}

\noindent where $N_k$ and $\lVert c_n-c_b^k \lVert$ are the number of superpixels in the $k$-th cluster,  and the Euclidean distance between $c_n$ and $c_b^k$.

{$\bullet$} \textbf{Knowledge Map from Traditional Methods}
We choose the saliency map from the contour-closure-based model (CCM) as one representative option for the external knowledge map. The CCM~\cite{liu2017hierarchical} highlights the importance of contour closure for salient object detection and propose to combine the closure completeness and the closure reliability for salient object detection. This CCM model proves to outperform the state-of-the-art unsupervised saliency models. Thus, the saliency map from CCM holds strong external knowledge to rectify inferior saliency information from traditional unsupervised candidate saliency models.

{$\bullet$} \textbf{Knowledge Map from Deep Methods}
If the deep saliency models are involved as candidates, the external knowledge based on traditional methods or assumptions will become relatively inferior. In such case, deep learning based external knowledge can be a preferred choice as the knowledge map. This paper introduces the resulted saliency map from the DHSNet model~\cite{liu2016dhsnet}. The DHSNet model utilizes a novel end-to-end deep hierarchical network based on convolutional neural networks for detecting salient objects. Evaluations prove that the DHSNet model shows significant superiority in terms of performance. Thus, we choose the saliency maps from DHSNet model as the external knowledge map if candidates involve deep models with extremely high performance.


\subsubsection{Consensus}
Even though an external knowledge map is introduced, its accuracy in saliency prediction can not be guaranteed just as the uncertainty in the candidate saliency models. Thus, we introduce a strict consistency scheme to reach a prudent consensus. In order to make a consensus, the arbitrator model judges the superpixel as salient only if the majority of the candidate saliency models vote it as salient as well as the external knowledge confirms its saliency. This consensus largely reduces the chances that an unsalient superpixel being misjudged as salient.

Given $P$ saliency maps, $S_{p}(n)$ is defined as the mean intensity value of the $n$-th superpixel on the $p$-th saliency map. The majority voting map is computed as

\begin{equation} \label{eq:maj}
    S_{\textup{Maj}}(n)=
\begin{cases}
    1,& \sum_{p = 1}^{P} \iota_{p,n}>\frac{P}{2}\\
    0, &  \text{otherwise}
\end{cases}
\end{equation}

A consensus map $S_{\textup{Con}}$ is computed by hearing voices from both the majority voting map and the external knowledge map:

\begin{equation} \label{eq:strictprior}
S_{\textup{Con}} = S_{\textup{Ext}} \times S_{\textup{Maj}},
\end{equation}

The multiplication operation in Eq.~\ref{eq:strictprior} makes the consensus map $S_{\textup{Con}}$ constrained by the majority voting map $S_{\textup{Maj}}$ and the external knowledge map $S_{\textup{Ext}}$. It largely reduces the numbers of false positives on these two maps.

\subsubsection{The Reference Map via Propagation}
The consensus map $S_{\textup{Con}}$ is a saliency map of high precision but only holds saliency information for certain parts of the image, so that we need to expand the saliency information to the whole image. A propagation method is employed to diffuse these saliency values on the consensus map to the whole image iteratively.

Propagation method firstly over-segments the image into superpixels and constructs an undirected graph, which comprises of a set of vertices of the superpixels together with a set of edges representing the similarity between adjacent vertices. Then, the propagation seeds~\cite{gopalakrishnan2010random,jiang2013saliency,ren2010improved,yang2013saliency,gong2015saliency} are selected to spatially diffuse to the whole graph within several iterations.

The over-segmented image can be regarded as an undirected graph $G = (V, E)$, which comprises a set $V$ of the superpixels together with a set $E$ of edges representing the similarity between adjacent superpixels. The constructed graph $G$ can be described as an adjacent matrix $W = {\left[ w_{nm}\right]}_{N\times N}$ with the similarity between two superpixels $c_n$ and $c_m$ computed as $w_{nm} = exp(- \mathcal{G}(c_n,c_m)^2/(2 \theta^2))$, where $\theta$ is set as 0.25 and $\mathcal{G}(c_n,c_m)$ computes the geodesic distance between $c_n$ and $c_m$:

\begin{equation} \label{eq:geodist}
\mathcal{G}(c_n,c_m) = \min_{V_1=n,V_2,\ldots,V_r=m} \left[ \sum_{k=1}^{r-1} {\max (\lVert V_{k}-V_{k+1} \lVert -a,0)}\right]
\end{equation}

\noindent \emph{s.t.} $V_k, V_{k+1} \in V$, $\lVert V_{k}-V_{k+1} \lVert$ computes the Euclidean distance between $V_{k}$ and $V_{k+1}$, and $a$ is an adaptive threshold preventing the ``small-weight-accumulation" problem~\cite{wei2012geodesic,gong2015saliency}. The $\mathcal{G}(c_n,c_m)$ measures the shortest path between $c_n$ and $c_m$ in the graph $G$.

Finally, we use a propagation function as follow to compute the reference map:
\begin{equation} \label{eq:propogation}
S^{t+1} = I\cdot D^{-1}\cdot W\cdot S^{t}.
\end{equation}
\noindent where $I$ is the identity matrix and $D$ is the diagonal degree matrix with $D_{nm} = \sum\nolimits_{m} w_{nm}$, the initial $S^{t=0}=S_{\textup{Con}}$, and after several times of iterations, the final propagated map is computed as the reference map $S_{\textup{Ref}}^0$. In practice, we set the propagation number as 5.

In this section, we propagate a reference map by taking the consensus of the external knowledge and the majority voting of all the candidate saliency models into consideration. The reference map integrated with the external knowledge is regarded as the reference map $S_{\textup{Ref}}^0$. Afterwards, each candidate saliency map is updated based on Eq.~\ref{eq:updating_rule}. Accordingly, the reference map $S_{\textup{Ref}}^t$ at $t>0$ is updated by averaging the candidate saliency maps as in Eq.~\ref{eq:sprior}. Thus, during the CA updating process, the reference map is updated by using the candidate saliency maps as in Eq.~\ref{eq:sprior} and the external knowledge is not integrated any more. The influence of the external knowledge is thus appropriate and recessive.

%% file: Expertise.tex
\section{Model-expertise Estimator}
\label{sec:expertise}
The $\alpha_p$ and $\beta_p$ measure the expertise of the $p$-th saliency model. $\alpha_p$ represents the expertise of the $p$-th saliency intensity map, which is a map with continuous values in the range of $[0,1]$. $\beta_p$ represents the expertise of the $p$-th binary map, which is a map with binary values $\{0,1\}$.

The saliency maps integrated into the framework involve both the intensity maps and the binary maps. In this work, the cellular automata (CA) is used to iteratively update the current saliency intensity map by involving the contextual influences of its neighborhood (other candidate saliency maps). Firstly, it is natural to design the candidate saliency map as an intensity map of continuous values for finer estimation, as it reflects the actual saliency intensity by the corresponding candidate saliency model. However, as intensity maps from various saliency models have diverse semantic meanings and magnitudes, we also adopt their corresponding binary maps as the contextual neighborhood to eliminate these two influences. It is convenient to transform an intensity map to a binary saliency map when necessary by introducing a threshold (like Otsu [45]). Otherwise, if it was the binary map to be updated in CA, it would cause inevitably information loss at the very beginning of the integration, and it would be impossible to recover an intensity one. Therefore, in the framework, both saliency intensity maps and saliency binary maps are integrated, but the current saliency maps to be updated in CA are intensity maps.

In this paper, we propose two online approaches to obtain the expertise of saliency models without supervised information from the image dataset, one is a statistics-based method from the intrinsic implications of Eq.~\ref{eq:ln_post_ratio}, the other is a latent-variable-based method for evaluating multiple models.

\subsection{Statistics-based Expertise}
\label{subsec:expertisep}
According to the Bayesian framework of Eq.~\ref{eq:ln_post_ratio}, we propose a statistics-based method to evaluate the expertise $\alpha_p$ and $\beta_p$. The statistics-based method analyzes the probability distributions of foreground and background samples on saliency maps and statistically computes $\alpha_p$ and $\beta_p$.

$\beta_p$ is the expertise of the $p$-th binary saliency map, which is originally derived from $\frac{P\left (  \boldsymbol{\iota}_{p} | F \right )}{P\left (  \boldsymbol{\iota}_{p} | \bar{F}  \right )}$. More specifically, $P\left (  \boldsymbol{\iota}_{p} | F \right )$ is ${P\left ( {\iota}_{p,n}=1 | F_{n} \right )}$, indicating the probability that the $n$-th superpixel on the $p$-th saliency map is labeled as foreground given the superpixel is a foreground one. Similarly, $P\left (  \boldsymbol{\iota}_{p} | \bar{F} \right )$ is ${P\left ( {\iota}_{p,n}=1 | \bar{F}_{n} \right )}$, indicating the probability that the $n$-th superpixel is miss-labeled as foreground given the superpixel is a background one. Although it is impossible to get the ground-truth $F$ in online methods, the reference map obtained in Section~\ref{sec:prior} can be regarded as the `best current' knowledge to approximate $F$.

In this work, as the burden of computing every local $\beta_{p,n}$ is rather heavy, we set a threshold $\lambda$ to classify the reference map $P(F)$ as foreground or background samples, and estimate a global $\beta_{p}$ to approximate the expertise of all the superpixels on the $p$-th saliency map. Then, the computation of $\beta_{p}$ is simplified as follows:

\begin{equation} \label{eq:prob_base}
\beta_{p} = \frac{P\left (  \boldsymbol{\iota}_{p} | F \right )}{P\left (  \boldsymbol{\iota}_{p} | \bar{F}  \right )}
= \frac{P\left ( {\iota}_{p,n}=1 | F_{n} \right )}{P\left ( {\iota}_{p,n}=1 | \bar{F}_{n} \right )}
\propto \frac{P\left ( {\iota}_{p}=1 | F \right )}{P\left ( {\iota}_{p}=1 | \bar{F} \right )}
\end{equation}

Thus, ${P\left ( {\iota}_{p}=1 | F \right )}$ and ${P\left ( {\iota}_{p}=1 | \bar{F} \right )}$ can be obtained only by their intrinsic implications of probability theory, namely

\begin{equation} \label{eq:iota_F}
P\left ( {\iota}_{p}=1 | F \right ) = \frac{P\left ( {\iota}_{p}=1, F \right )}{P\left (F \right )},
\end{equation}

\begin{equation} \label{eq:iota_F_bar}
P\left ( {\iota}_{p}=1 | \bar{F} \right ) = \frac{P\left ( {\iota}_{p}=1, \bar{F} \right )}{P\left (\bar{F} \right )},
\end{equation}

More specifically, the probability functions $P\left ( {\iota}_{p}=1, F \right )$ and $P\left ( {\iota}_{p}=1, \bar{F} \right )$ are statistically computed as follow:

\footnotesize
\begin{equation} \label{eq:prob_iota_F}
P\left ( {\iota}_{p}=1, F \right ) = \frac{1}{N}{\sum_{n=1}^{N} [\mathcal{T}(S_{p}(n), \gamma_{p})\cdot \mathcal{T}(S_{\textup{Ref}}(n), \lambda)]},
\end{equation}

\begin{equation} \label{eq:prob_iota_F_bar}
P\left ( {\iota}_{p}=1, \bar{F} \right ) = \frac{1}{N} {\sum_{n=1}^{N} [\mathcal{T}(S_{p}(n), \gamma_{p})\cdot \left (1-\mathcal{T}(S_{\textup{Ref}}(n), \lambda)\right )]},
\end{equation}
\normalsize

\noindent where $N$ is the number of superpixels on the over-segmented image. $S_{p}(n)$ is the mean intensity value of the $n$-th superpixel on the $p$-th saliency intensity map, and $\gamma_{p}$ is the OTSU threshold~\cite{otsu1975threshold} of the $p$-th saliency intensity map. $S_{\textup{Ref}}(n)$ is the mean intensity value of the $n$-th superpixel of the reference map. $\mathcal{T}$ is a thresholding function as follow:

\begin{equation} \label{eq:thresh}
    \mathcal{T}(\mu, \nu) =
\begin{cases}
    1,& \mu \geq \nu\\
    0, &  \text{otherwise},
\end{cases}
\end{equation}

$P\left (F \right )$ is computed as
\begin{equation} \label{eq:prob_prior}
P\left (F \right ) = \frac{1}{N}\sum_{n=1}^{N} \mathcal{T}(S_{\textup{Ref}}(n), \lambda),
\end{equation}

\noindent and $P\left (\bar{F} \right ) = 1-P\left (F \right )$. Thus, a global $\beta_{p}$ of the $p$-th binary saliency map is computed based on probability theory.

$\alpha_p$ represents the expertise of the $p$-th saliency intensity
map, which can be computed in a similar way as computing $\beta_p$. Thus, we have

\begin{equation} \label{eq:prob_base_s}
\alpha_{p} = \frac{P\left ( {s}_{p} | F \right )}{P\left (  {s}_{p} | \bar{F}  \right )}= \frac{\frac{P(s_p,F)}{P(F)}}{\frac{P(s_p,\bar{F})}{P(\bar{F})}}
\end{equation}

However, as $s_p$ is a map with continuous values other than discrete ones, we employ a fixed stepsize of 0.1 in $[0.1, 0.9]$ to binarize $s_p$ with gradually increasing thresholds, and $\alpha_p$ is the mean ratio of $\frac{P\left (  \boldsymbol{\iota}_{p} | F \right )}{P\left (  \boldsymbol{\iota}_{p} | \bar{F}  \right )}$ with all the thresholds.

We denote the step-sized thresholds as a set of $\lambda^{'}=\{0.1, 0.2, 0.3, \ldots, 0.9\}$, where $J=9$ is the number of thresholds in the set $\lambda^{'}$. Then the probability functions $P\left ( s_p, F \right )$ and $P\left ( s_p, \bar{F}  \right )$ are statistically computed as

\begin{equation} \label{eq:s_prob_iota_F}
\begin{split}
&P\left ( {s}_{p}, F \right )\\
&= \frac{1}{N} \frac{1}{J} \sum_{j=1}^{J} \sum_{n=1}^{N} [\mathcal{T}(S_{p}(n), \gamma_{p})\cdot \mathcal{T}(S_{\textup{Ref}}(n), \lambda^{'}(j))],
\end{split}
\end{equation}

\begin{equation} \label{eq:s_prob_iota_F_bar}
\begin{split}
&P\left ( {s}_{p}, \bar{F} \right )\\
&= \frac{1}{N} \frac{1}{J} \sum_{j=1}^{J} \sum_{n=1}^{N} [\mathcal{T}(S_{p}(n), \gamma_{p})\cdot (1-\mathcal{T}(S_{\textup{Ref}}(n), \lambda^{'}(j)))].
\end{split}
\end{equation}
\normalsize

With the probability theory, we finally compute the expertise of the $p$-th saliency intensity map $\alpha_{p}$ and the expertise of the $p$-th binary saliency map $\beta_{p}$ in a statistical way.

\input{tab/singlemodel_mae.tex}

\subsection{Latent-variable-based Expertise}
The candidate saliency models distinguish a superpixel on an image as salient or not based on the corresponding saliency maps. Besides, each superpixel on the image is assumed to possess difficulty for saliency assessment, namely $\pi_{n}$. In recent saliency integration approaches~\cite{zhang2017supervision,Quan2017Unsupervised}, the concept of superpixel difficulty are adopted in the process of computing the expertise of the candidate saliency map. The expertise $\beta_{p}$ as well as the difficulty of the superpixel $\pi_{n}$ are assumed as latent variables and are solved by optimizations.

$\beta_p$ represents the expertise of the $p$-th binary saliency map, which is assumed to range $\beta_p \in (-\infty, +\infty)$. If $\beta_p<0$, the $p$-th candidate model makes wrong measurements and shows inferior ability in saliency prediction. If $\beta_p>0$, the $p$-th model makes correct measurements and shows superior ability in saliency prediction. When $\beta_p=0$, the $p$-th model is not able to distinguish saliency objects. $\beta_p = +\infty$ implicates that the $p$-th model always makes correct decisions about saliency objects, while $\beta_p = -\infty$ means that the $p$-th binary saliency map always misjudge saliency information.

Besides the assumption that candidate models vary in expertise $\beta_p$, we presume that each superpixel in an image has varying degrees of difficulty for saliency assessment and introduce a measurement $\pi_{n}\in[0,+\infty)$ to represent the difficulty of a superpixel. $\pi_n=0$ means that the superpixel possesses extremely low difficulty such that even an inexperienced saliency model can distinguish its saliency. On the contrary, $\pi_n=+\infty$ means the superpixel is so ambiguous that even the best saliency model has a chance to misjudge it.

As defined in Section~\ref{subsec:framework}, $l_n$ is the true binary saliency label of the $n$-th superpixel on the given image, while $\iota_{p,n}$ refers to the actual binary saliency label of the $n$-th superpixel by the $p$-th model. Thus, the probability that the $p$-th model correctly labels a superpixel on an image is

\begin{equation} \label{eq:prob_labelcorrect}
    p\left ( \iota_{p,n} = l_n|\beta_{p},\pi_n\right ) =
\begin{cases}
    1,& \pi_n = 0\\
    \frac{1}{1+e^{-\beta_{p}/ \pi_n}}, &  \text{otherwise}
\end{cases}
\end{equation}

More skilled saliency models (higher $\beta_p$) have a higher probability of correctly labeling a superpixel. As the difficulty $\pi_{n}$ of a superpixel increases, the probability of correctly labeling the superpixel decreases, and vice versa.

Now, given a set of actual saliency labels by multiple saliency models $\boldsymbol{\iota}=\{\iota_{p,n}\}$, our goal is to estimate the unobserved latent parameters including the true saliency labels of superpixels $\boldsymbol{l}=\{l_ {n}\}$, the expertise of the candidate models $\boldsymbol{\beta}=\{\beta_p\}$, and the difficulties of the superpixels $\boldsymbol{\pi}=\{\pi_n\}$. Here the Expectation-Maximization (EM) algorithm is used to achieve the optimal values of the latent parameters.

In the E-step, we compute the posterior probabilities of $l_n$ with the parameters $\boldsymbol{\beta}$, $\boldsymbol{\pi}$ obtained from the last M-step and the actual labels:

\begin{equation} \label{eq:e_step}
\begin{split}
p(l_n|\boldsymbol{\iota},\boldsymbol{\beta},\boldsymbol{\pi}) &= p(l_n|\boldsymbol{\iota}_n,\boldsymbol{\beta},\pi_n) \\
 &\propto p(l_n|\boldsymbol{\beta},\pi_n)p(\boldsymbol{\iota}_n|l_n,\boldsymbol{\beta},\pi_n) \\
 &\propto p(l_n)\prod_p (\iota_{p,n}|l_n,\beta_p,\pi_n),
\end{split}
\end{equation}

\noindent where $\boldsymbol{\iota}_n$ denotes the actual labels of a superpixel by all the $P$ candidate models and the parameters $\boldsymbol{\beta}$, $\boldsymbol{\pi}$ are conditionally independent of $l_n$. In practice, we use Gaussian distribution $(\mu=\theta=1)$ for $\beta$, re-sample $1/\boldsymbol{\pi}$ as $e^{(1/\boldsymbol{\pi}')}$, and use the same Gaussian distribution on $1/\boldsymbol{\pi}'$ to avoid $\boldsymbol{\pi}$ being negative.

In the M-step, we compute the expected value of the log likelihood function with respect to the conditional distribution of $\boldsymbol{l}$ given $\boldsymbol{\iota}$ under the current estimate of $\boldsymbol{\beta}$ and $\boldsymbol{\pi}$ as follows:

\begin{equation} \label{eq:m_step}
\begin{split}
Q&\left ( \boldsymbol{\beta},\boldsymbol{\pi} \right )
=E\left [ \ln p\left ( \boldsymbol{\iota},\mathbf{l}|\boldsymbol{\beta},\boldsymbol{\pi} \right ) \right ]\\
&=E\left [ \ln\prod_{n}\left ( p\left ( l_n \right ) \prod_{p} p\left ( \iota_{p,n}|l_n, \beta_p,\pi_n \right )\right ) \right ]\\
&=\sum_{n}E\left [ \ln p\left ( l_n \right )\right ] + \sum_{p,n} E\left [\ln p\left ( \iota_{p,n}|l_n, \beta_p,\pi_n \right ) \right ],
\end{split}
\end{equation}
since $\iota_{p,n}$ are conditionally independent given $\boldsymbol{l}$, $\boldsymbol{\beta}$, $\boldsymbol{\pi}$.
With gradient ascent method, the parameters $\boldsymbol{\beta}$ and $\boldsymbol{\pi}$ are set to maximize the quantity function $Q$ in Eq.~\ref{eq:m_step}.

Here, we presume that the expertise of the $p$-th saliency intensity map $\alpha_{p}$ is equal to $\beta_{p}$ to simplify the computation. The details of the EM algorithm can be found in~\cite{whitehill2009whose}.

%% file: tab/singlemodel_mae.tex
\begin{table*}[t]
\begin{center}
    \small
    \begin{tabular}{|c|c|c|c|c|c|c|c|c|c|c|c|}
    \hline
    Model&DRFI	&MB+	&RB	&TLLT	&MB	&BSCA	&RC	&MR	&GP	&UFO	&COV\\\hline
    MAE&0.170	&0.171	&0.172	&0.172	&0.174	&0.183	&0.186	&0.186	&0.191	&0.203	&0.220\\
    F-measure&0.765&0.722&0.710&0.717&0.709&0.736&0.720&0.735&0.725&0.684&0.602\\
    \hline\hline
    Model   &HS	&GC	&CEOS	&PCAS	&GBVS	&LR	&IT	&FT	&CA	&SR	&IS\\\hline
    MAE&0.228	&0.234	&0.243	&0.247	&0.263	&0.274	&0.289	&0.291	&0.309	&0.311	&0.334\\
    F-measure&0.700&0.564&0.646&0.622&0.599&0.622&0.520&0.384&0.483&0.409&0.416\\
    \hline
    \end{tabular}\\[0.5ex]

    \begin{tabular}{|c|c|c|c|c|c|c|c|}
    \hline
    Model&DHSNet	&DSS	&DCL	&MDF	&RFCN   &CCM	&Bound\\\hline
    MAE&0.059	&0.062	&0.068	&0.135	&0.147	&0.151	&0.237\\
    F-measure&0.886	&0.884	&0.888	&0.796	&0.878	&0.738	&0.581\\
    \hline
    \end{tabular}\\[1ex]
      \caption{The list of the twenty-seven candidate saliency models. The models are ranked by their MAE evaluation results on the ECSSD dataset and their mean F-measure scores (with adaptive thresholds) are also reported. In the bottom part, the performances of deep models, CCM model and the pre-computed boundary-prior-based external knowledge map are presented.}
    \label{tbl:rank}
    \end{center}
\end{table*} 

%% file: experiments.tex
\begin{figure*}
\begin{center}
\begin{minipage}{0.3\linewidth}
    \centerline{\includegraphics[width=\linewidth]{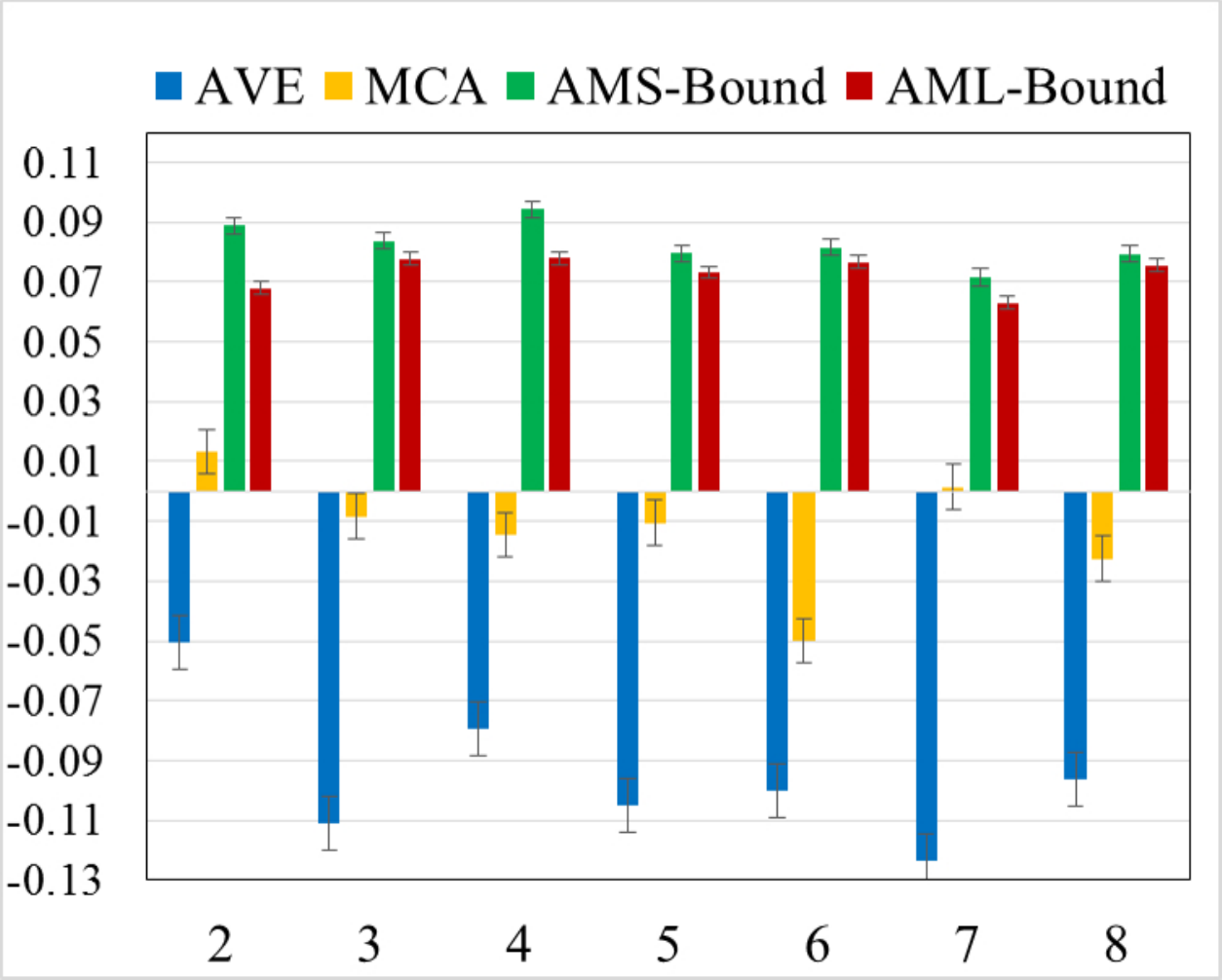}}
    \centerline{(a) F-measure}\medskip
\end{minipage}
\begin{minipage}{0.3\linewidth}
     \centerline{\includegraphics[width=\linewidth]{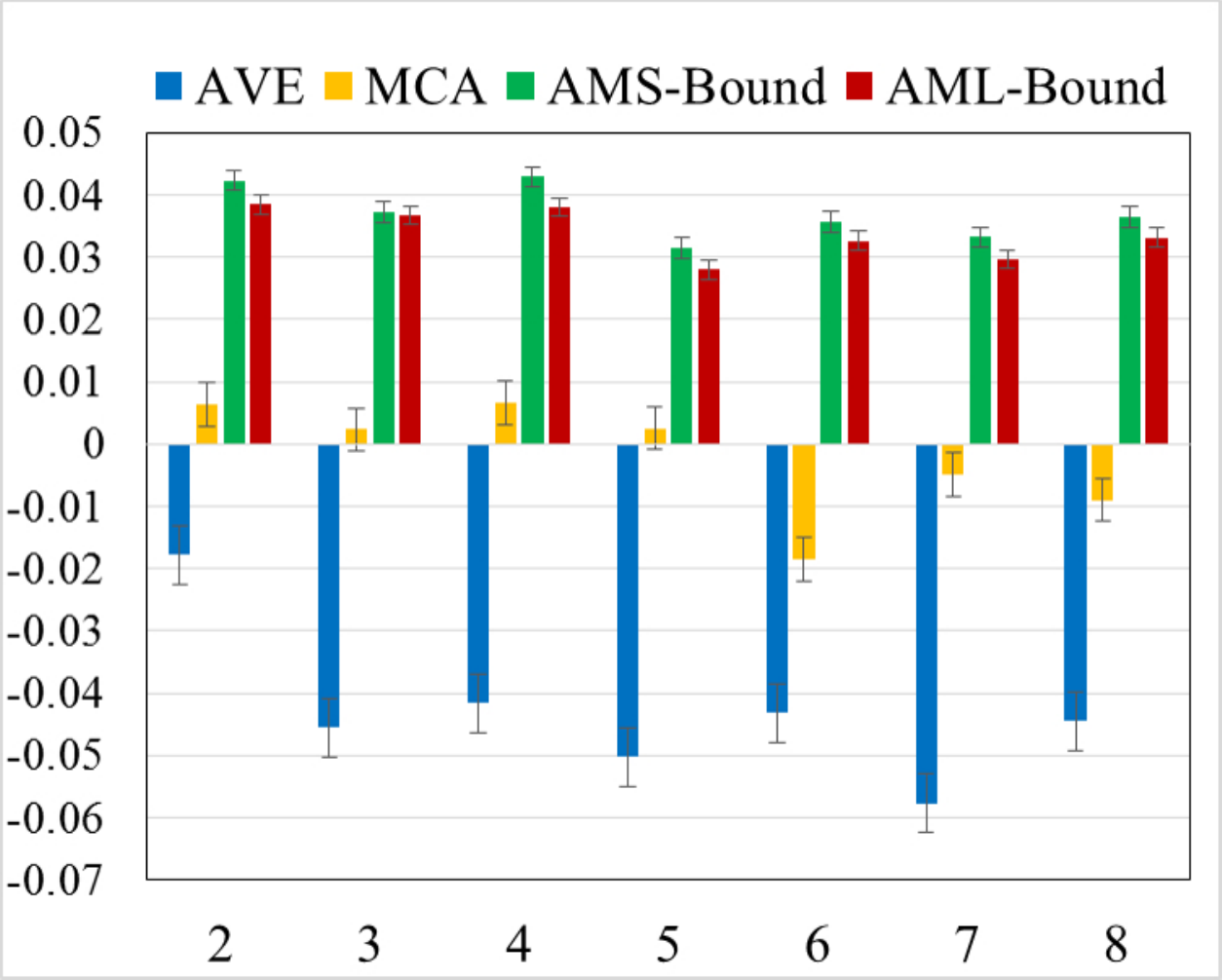}}
     \centerline{(b) MAE}\medskip
\end{minipage}
\begin{minipage}{0.3\linewidth}
    \centerline{\includegraphics[width=\linewidth]{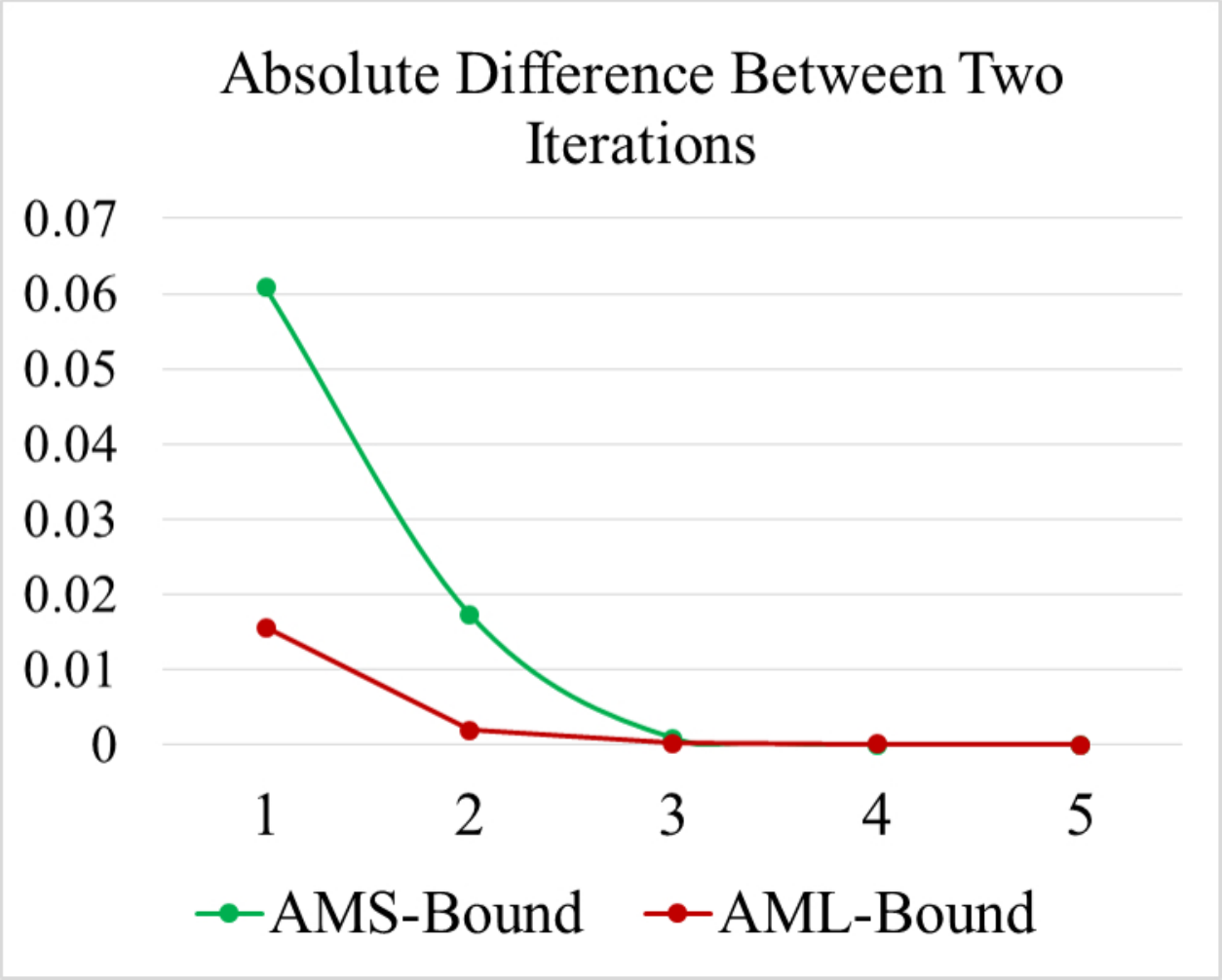}}
    \centerline{(c) Convergence Experiment}\medskip
\end{minipage}
\caption{(a)-(b) The average performance enhancement of five randomly selected combinations for each fixed number combination using strategy 3 in Section~\ref{subsec:ecssd} compared to their corresponding top models. (a) measures the mean F-measure improvement. (b) measures the average improvement of MAE scores. The average maps (AVE), MCA, AMS and AML results are compared and horizontal axis indicates the number of candidate models being combined. (c) Convergence experiments computing the average absolute difference of all superpixels between two generations.}
\label{fig:fm}
\end{center}
\end{figure*}

\section{Experiments}
\label{sec:experiment}
The arbitrator model (AM) aims at generating a saliency integration model that solidly enhances the performance regardless of the choices of candidate saliency models. Any saliency models can be selected for saliency integration in AM and no special assumptions on saliency models are required.

In this section, we perform a comprehensive evaluation of the AM model under various combination strategies by adopting the state-of-the-art saliency models as the candidates. We choose twenty-seven state-of-the-art saliency models including the traditional models BSCA~\cite{qin2015saliency}, CA~\cite{goferman2012context}, CEOS~\cite{mairon2014closer}, COV~\cite{erdem2013visual}, DRFI~\cite{jiang2013salient}, FT~\cite{achanta2009frequency}, GBVS~\cite{harel2006graph}, GC~\cite{cheng2015global}, GP~\cite{jiang2015generic}, HS~\cite{yan2013hierarchical}, IS~\cite{hou2012image}, IT~\cite{itti1998model}, LR~\cite{shen2012unified}, MB~\cite{zhang2015minimum}, MB+~\cite{zhang2015minimum}, MR~\cite{yang2013saliency}, PCAS~\cite{margolin2013makes}, RB~\cite{wei2012geodesic}, RC~\cite{cheng2015global}, SR~\cite{hou2007saliency}, TLLT~\cite{gong2015saliency}, UFO~\cite{jiang2013salient}, and deep models including DSS~\cite{hou2017saliency}, DCL~\cite{li2016deep}, RFCN~\cite{dai2016r}, MDF~\cite{li2015visual}, and DHSNet~\cite{liu2016dhsnet}. The implementations of the chosen approaches are directly from the corresponding authors.

For comprehensive evaluation, four challenging datasets are utilized in the experiments: ECSSD~\cite{yan2013hierarchical}, ASD~\cite{achanta2009frequency}, ImgSal~\cite{li2013visual} and DUT-OMRON~\cite{yang2013saliency}. The ASD dataset is one of the most widely used datasets with 1000 images from the MSRA-5000 Saliency Object Database\cite{liu2011learning}, with distinct salient objects on the scenes. The ImgSal dataset is challenging, including 235 images in six levels of complexity. The ECSSD dataset contains 1000 images with complex salient objects on the scenes, and the objects on the images are semantically meaningful. The DUT-OMRON dataset contains a large number of 5168 more difficult and challenging images.

\subsection{Implementation and Evaluation}
We over-segment the images into $N = 400$ superpixels with the simple linear iterative clustering (SLIC) algorithm~\cite{achanta2012slic}. In practice, we set the numbers of generations in Eq.~\ref{eq:final_saliency} as 5 for the CA updates, and $\lambda$ is set as 0.1. We reference our arbitrator model with statistics-based expertise as AMS and with latent-variable-based expertise as AML in all the experiments. Moreover, we refer ``-B'', ``-C'', ``-D'' as the boundary-based external knowledge, contour-closure-based external knowledge and deep-based external knowledge respectively (\emph{i.e.}, AMS-B means that the AM model uses the boundary-based external knowledge in the reference generator and adopts the statistics-based expertise estimator). Besides the saliency maps computed from AMS and AML, we also compute the average saliency maps of the candidate saliency models (AVE), BN~\cite{borji2012salient}, M-estimator~\cite{le2014saliency} and MCA~\cite{qin2015saliency} for fair comparisons. All the existing saliency integration models being selected in this work are online models, and the codes are provided by the corresponding authors with recommended parameter settings.

We employ two types of evaluation metrics to evaluate the performance of saliency maps: F-measure and mean absolute error (MAE). F-measure is computed to count for the saliency maps with both high precision and recall:

\begin{equation} \label{eq:fmeasure}
F = \frac{\left(1+{\beta}^2\right)\cdot precision\cdot recall}{\beta^2 \cdot precision + recall},
\end{equation}
where $\beta^2 = 0.3$~\cite{achanta2009frequency} to emphasize the precision, and the precision and recall are obtained by using twice the mean saliency values of the saliency maps as adaptive thresholds~\cite{achanta2009frequency}.

MAE measures the overall pixel-wise difference between the saliency map $sal$ and the ground truth $gt$: 

\begin{equation} \label{eq:mae}
MAE = \frac{1}{H}\sum_{h=1}^{H} {\left|sal(h)-gt(h)\right|}.
\end{equation}

\noindent where $H$ is the number of pixels on the map.


\input{tab/ecssddataset_adpt.tex}

\input{tab/5dataset.tex}

\subsection{Comparisons of Various Combinations}
\label{subsec:ecssd}
We choose the number of candidate saliency models for integration from 2 to 8. If enumerating all the possible combinations from 2 to 8 models, we need to evaluate $C_{27}^2+C_{27}^3 \ldots + C_{27}^8 = 3,505,671$ combinations, which is almost impossible. Thus, we follow four different strategies to evaluate fifty-eight representative combinations. Table~\ref{tbl:rank} lists the performances of the twenty-seven candidate saliency models on ECSSD dataset by ranking the MAE.

In Table~\ref{tbl:ecssd}, we list the mean F-measure of the proposed AM model with four different combination strategies. We compare the integrated results of the AM model with every candidate saliency model being combined (only list the one with the best performance in column ``Top'' and refer it as top model in experimental analysis), the average saliency maps (AVE), the resulted BN, M-estimator (M-est), and MCA saliency maps. The detailed evaluation and analysis of the four strategies are listed below.


1. \textbf{Superior models combination.}
When choosing the candidate saliency models, we only consider those saliency models with the best performances. Thus, we choose two best saliency models for 2-model-combination, three best saliency models for 3-model-combination and so forth. The first seven rows in Table~\ref{tbl:ecssd} indicate the evaluation results, where it can be easily perceived that both AMS and AML outperform the top candidate saliency model as well as the MCA model in every combination. Thus, the proposed AM model performs well when superior saliency models are combined.

2. \textbf{Inferior models combination.} We only consider those saliency models with the worst performances. For example, we choose the worst two saliency models for 2-model-combination, the worst three saliency models for 3-model-combination and so forth. Table~\ref{tbl:ecssd} presents the evaluation results. Obviously, the AM model largely improves the F-measure of the top candidate saliency model with an average increase of 6.6\%, 11.0\%, 17.8\%, 7.4\%, 12.7\%, and 20.9\% for AMS-B, AMS-C, AMS-D, AML-B, AML-C, and AML-D correspondingly, while other online integration models such as AVE, BN, MCA and M-est decrease the F-measure by averagely 3.6\%, 3.6\%, 1.8\% and 3.6\% respectively. Apparently, the AM model greatly rectifies the error saliency information from the inferior candidate saliency models.

3. \textbf{Random combination.} From 2-model combination to 8-model combination, we randomly select candidate saliency models from the model pool and randomly evaluate five different combinations for each fixed number combination. The group ``Random Models Combinations'' in Table~\ref{tbl:ecssd} shows one example of each fixed number combination with random selection strategy. Again, the AM model consistently outperforms each one of the combined saliency models and the MCA model. Figure~\ref{fig:fm} indicates the performance enhancement of the average maps, MCA model, AMS and AML model compared to the corresponding top models by averaging five random combinations for each fixed number combination. Apparently, the proposed model solidly improves the performance independent of the number of models being chosen for combination.

\begin{figure*}
\begin{center}
\begin{minipage}{0.35\linewidth}
    \centerline{\includegraphics[width=\linewidth]{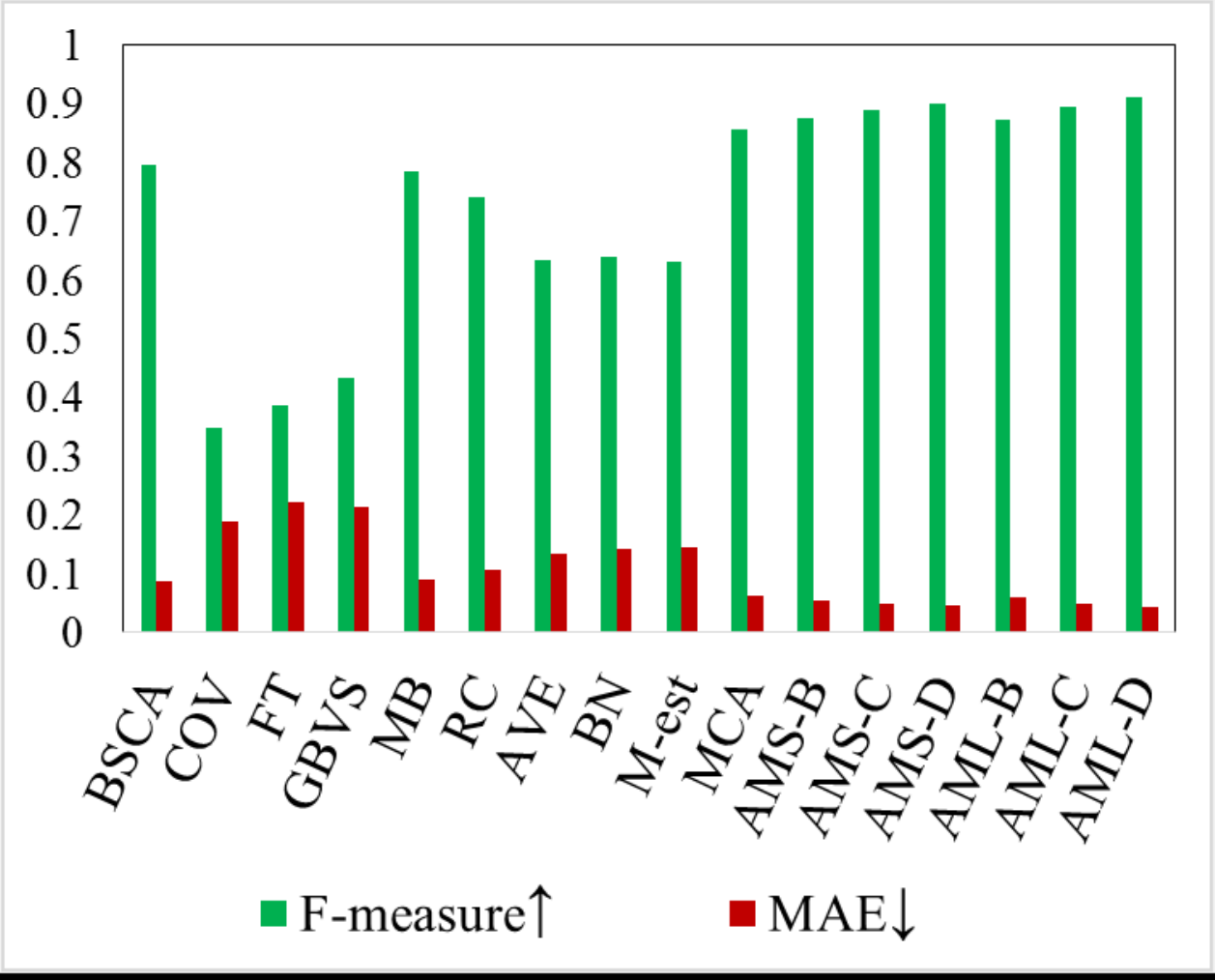}}
    \centerline{(a) ASD}\medskip
\end{minipage}
\begin{minipage}{0.35\linewidth}
    \centerline{ \includegraphics[width=\linewidth]{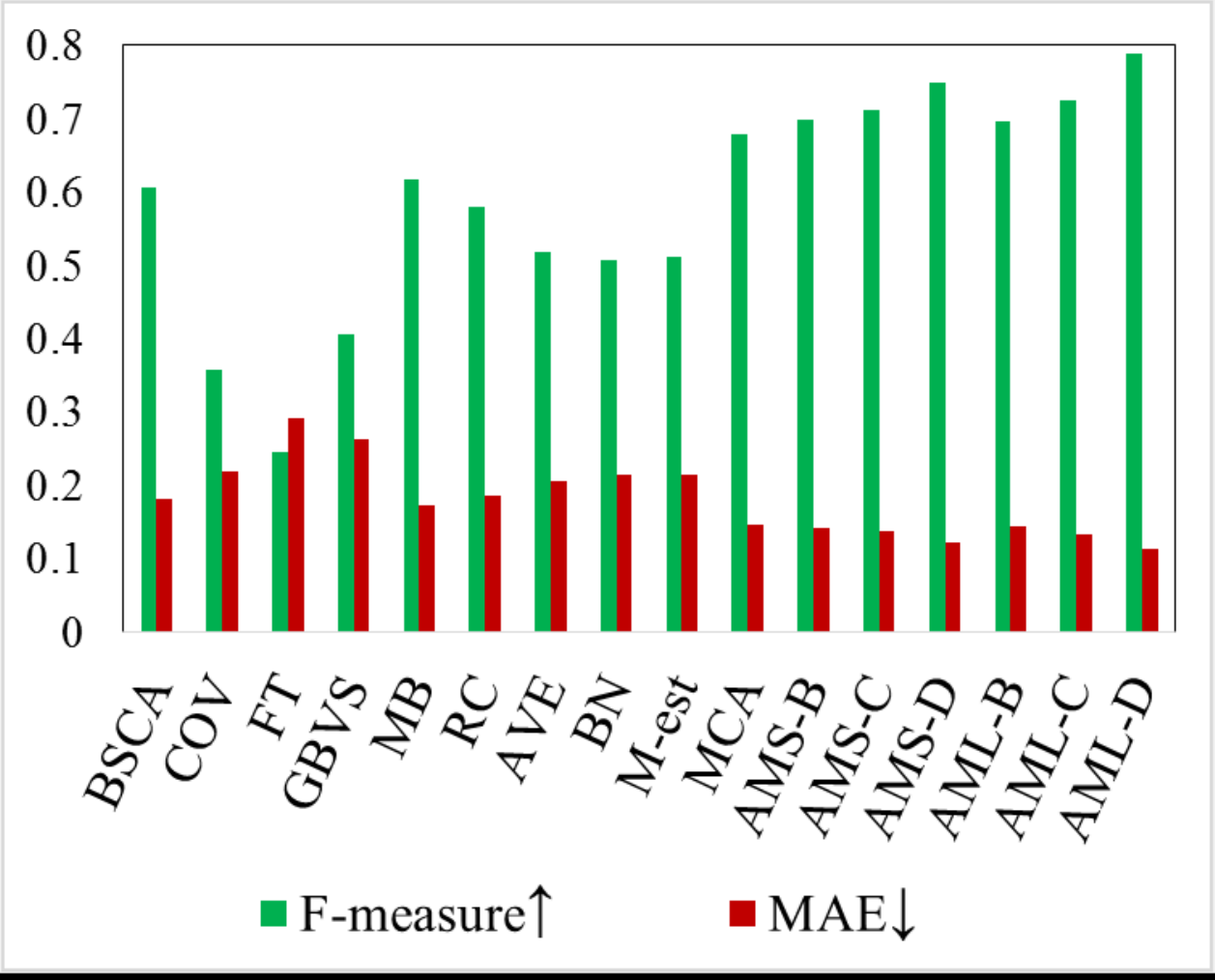}}
     \centerline{(b) ECSSD}\medskip
\end{minipage}
\\[1ex]
\begin{minipage}{0.35\linewidth}
     \centerline{\includegraphics[width=\linewidth]{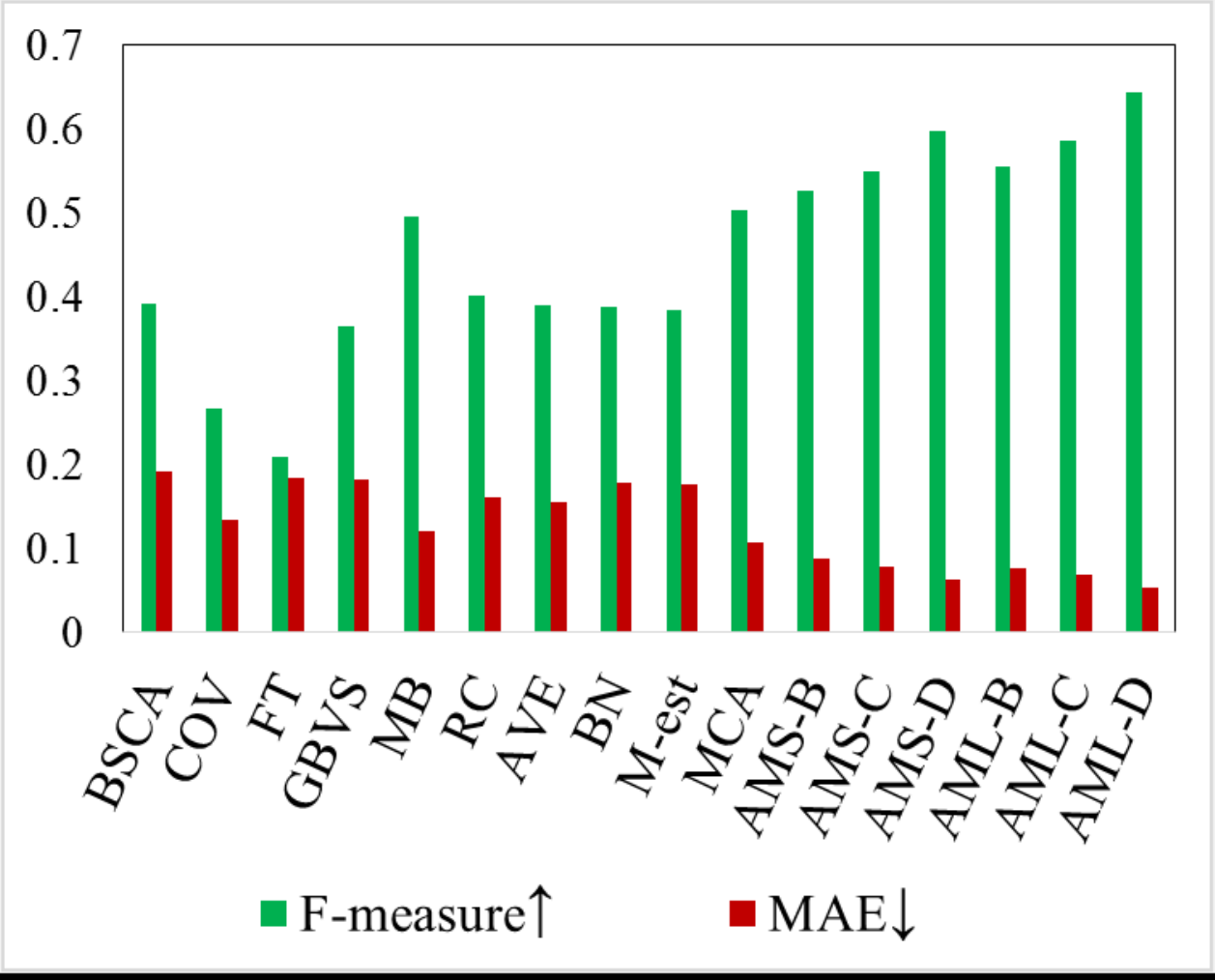}}
     \centerline{(c) ImgSal}\medskip
\end{minipage}
\begin{minipage}{0.35\linewidth}
    \centerline{ \includegraphics[width=\linewidth]{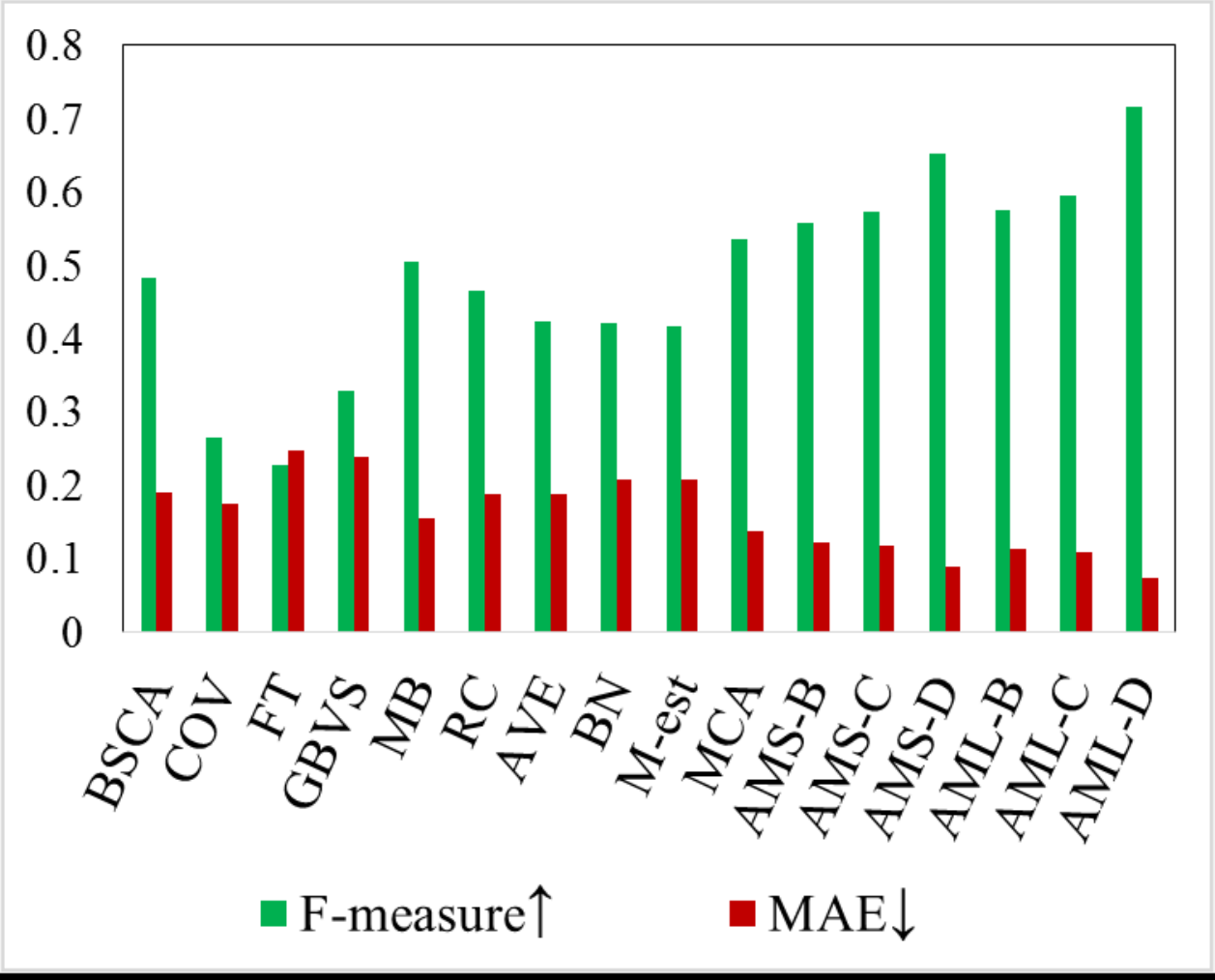}}
     \centerline{(d) DUT-OMRON}\medskip
\end{minipage}
\caption{(a)-(d) Average MAE and F-measure of candidate saliency models, average saliency maps (AVE), BN, M-estimator (M-est), MCA and the AM model on the four datasets including ECSSD, ASD, ImgSal, and DUT-OMRON. }
\label{fig:fivedataset}
\end{center}
\end{figure*}

Also, we evaluate our AM model over four challenging datasets, ECSSD~\cite{yan2013hierarchical}, ASD~\cite{achanta2009frequency}, ImgSal~\cite{li2013visual} and DUT-OMRON~\cite{yang2013saliency}. We use the combination of MB~\cite{zhang2015minimum}, BSCA~\cite{qin2015saliency}, RC~\cite{cheng2015global}, GBVS~\cite{harel2006graph}, COV~\cite{erdem2013visual} and FT~\cite{achanta2009frequency} as an example. Table~\ref{tbl:5dataset} presents the F-measure of the top candidate saliency models, average saliency maps (AVE), BN, M-estimator, MCA results and our AM results over the four datasets. Our AM model largely improves the performance compared to the best candidate model on all the four datasets, and outperforms the average maps and MCA model all the time. Figure~\ref{fig:fivedataset} (a)-(d) present the average MAE and F-measure of the candidate models being combined, the average saliency maps (AVE), results from BN, M-est and MCA model, and results from the AM model on the four datasets.

4. \textbf{Deep models combination.}
In the experiment, we select deep models including DSS~\cite{hou2017saliency}, DCL~\cite{li2016deep}, RFCN~\cite{dai2016r}, MDF~\cite{li2015visual}, and DHSNet~\cite{liu2016dhsnet} for evaluation. In the last group ``Deep Models Combinations'' in Table~\ref{tbl:ecssd}, we present different combinations involving deep models and traditional models. From the results, when deep models are involved as candidates, all the integration results of AM model that incorporate deep-network-based reference maps outperform the top candidate models. The first four rows are integration with all deep saliency models. The AMS-D and AML-D with deep external knowledge maps surpass the top saliency models averagely by 0.9\% and 1.4\% respectively. In the last four rows, the F-measure of the AVE, BN, MCA and M-est integration models drop sharply compared to the top models by averagely 5.3\%, 5.2\%, 9.7\%, and 4.6\% respectively, while AMS-D and AML-D averagely increase by 0.3\% and 0.2\% respectively.

In general, the AM model outperforms the existing integration models with all the four combination strategies. When the candidate saliency maps are from traditional models (\textit{i.e.}, ``Superior Models Combinations'', ``Inferior Models Combinations'' and ``Random Combinations''), AMS-B, AMS-C and AMS-D increase the F-measure of the top saliency models by averagely 3.1\%, 5.0\% and 10.0\% respectively, while AML-B, AML-C and AML-D increase the F-measure of the top models by averagely 3.2\%, 6.1\% and 12.5\% respectively. Thus, we conclude that the AM model solidly improves the performance regardless of the candidate models being combined.

\begin{figure*}
\begin{center}
\scalebox{1}{\includegraphics[width=7in]{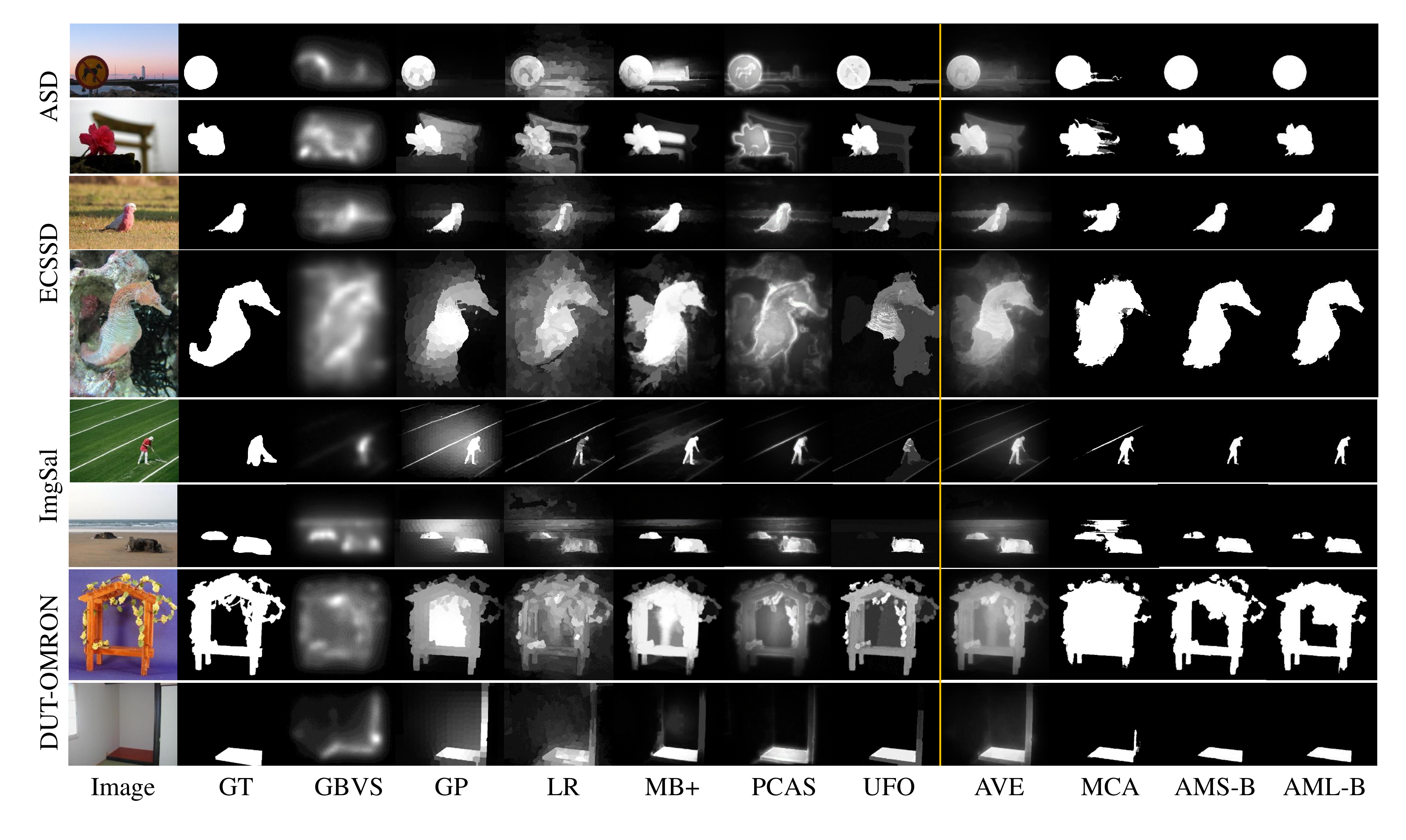}}
\vspace*{-5mm}
\caption{Examples of the results of combined saliency models, average saliency maps (AVE), BN, M-estimator (M-est), MCA, AMS and AML. The images with ground truth (GT) are sequentially from ECSSD, ASD, ImgSal, and DUT-OMRON datasets.}
\label{fig:example}
\end{center}
\end{figure*}

Figure~\ref{fig:example} shows some examples of the results of candidate saliency models, the average saliency maps, MCA model and the AM model on the four datasets.


\subsection{Rationality of the Reference Map and the Expertise}
By evaluating the AM model with various combinations following four strategies, we conclude that the AM model substantially improves the performance regardless of the candidate models. In this section, we discuss the rationality of the reference map and the expertise respectively.

As mentioned in Section~\ref{sec:prior}, the reference map $S_{\textup{Ref}}$ is directly derived from $P(F)$ in Eq.~\ref{eq:ln_post_ratio}, so that it should provide a natural scheme to introduce the information about salient object detection. In practice, the reference map is propagated from the consensus map of the external knowledge and the candidate saliency maps. Theoretically, the external knowledge can be any reasonable assumptions or currently existing models. In this paper, we address the inevitability of the reference map as one of the main components of the AM model. Thus, we firstly investigate the necessity of the reference map and then discuss the influences of different selections of the external knowledge.

We choose DRFI, GP, LR, MB+, TLLT and UFO as candidate saliency models and report the mean F-measure of the saliency integration results based on ECSSD dataset. As in Table~\ref{tbl:priorcandidate}, the first row shows the performances when different external knowledge is incorporated to compute the reference map, while the second row uses the external knowledge directly as an additional candidate saliency map. Obviously, the external knowledge being incorporated into the reference map results in better integration than it being integrated as another candidate map. Even when the external knowledge is a saliency model with inferior performance, incorporating the external knowledge for the reference map results in better integration results than taking the external knowledge as another candidate saliency model (first column of Table~\ref{tbl:priorcandidate}). Thus, introducing the external knowledge receives performance enhancement in practice.

Further, from Table~\ref{tbl:ecssd}, it can be perceived that when using the same expertise estimation method, the better the quality of the reference map, the higher performance the integration can be received. Apparently, the incorporation of the reference map is critical to the performance enhancement.

As mentioned in Section~\ref{sec:prior}, during the CA updating process, the reference map is updated by averaging the candidate saliency maps. Thus, we conduct a small scale experiment to evaluate whether the reference map should be updated at every iteration. We test with three combinations on ECSSD dataset: 1) DHSNet, GP, IT, 2) GP, LR, PCAS, and 3) GC, GP, PCAS. By updating the reference map in each generation, the mean F-measure of the above three combinations are higher than keeping the reference map unchanged by averagely 2.01\%, 0.23\% and 0.06\% for AMS-B, AMS-C, AMS-D and 1.30\%, 0.33\%, 0.07\% for AML-B, AML-C, AML-D respectively. Thus, we finally decide to update the reference map by considering both the theoretical inference and the practice.

According to Table~\ref{tbl:ecssd}, AML performs similarly well or slightly better than AMS in general when the same external knowledge is introduced. However, there are some exceptions. For instance, the AMS-B outperforms the AML-B by averagely 2.2\% in ``Deep Models Combinations''. The reason behind this is that different from the latent-variable-based expertise, the statistics-based expertise borrows information from the reference map as the approximated `Ground Truth' as in Section~\ref{subsec:expertisep}. Thus, the accuracy of the statistics-based expertise is influenced by the quality of the reference map. In ``Deep Models Combinations'', the high performances of deep candidates largely enhance the quality of the reference map, such that the AMS-B outperforms the AML-B significantly.

To quantitatively investigate the contributions of the reference map and the expertise, we also experiment each unit of the AM model on ECSSD dataset with a combination of candidate models including DRFI, GP, LR, MB+, TLLT and UFO, as shown in Table~\ref{tbl:priorexp}. The basic framework is an integration of candidate saliency models with equal expertise but without the reference map based on CA. Then we add different components to the basic framework, where the letter ``L'' refers to the latent-variable-based expertise, ``S'' means the statistics-based expertise, and ``Ref-B'' uses the boundary-prior-based external knowledge. From Table~\ref{tbl:priorexp}, it is obvious that the introduction of the reference map significantly enhances the performance of the basic CA framework. The involvement of the statistics-based expertise and the latent-variable-based expertise also improves the performance of the basic framework respectively. Finally, the statistics-based integration and the latent-variable-based integration with the reference map result in similarly better performances than only incorporating single unit in CA. Thus, the incorporation of both the reference map and the expertise of saliency models results in the best performance.

The proposed AM model synchronizes the $p$-th candidate by using its continuous map and the other candidates as binary maps during the CA process. Such updating form keeps the actual saliency intensity of the current candidate map and eliminates diverse semantic meanings and magnitudes from the other saliency models, as mentioned in Section IV. In Table~\ref{tbl:binary_continous}, we present the mean F-measure of the integration results by using 1) only continuous maps for integration, 2) only binary maps for integration, and 3) both continuous and binary maps for integration. It can be perceived that using both the continuous maps and binary maps produces the best integration results. Although the resulted maps by using both continuous and binary maps are only slightly better than or equal to the results by using only binary maps, it does not increase the computation complexity and even reduces the numbers of thresholding process once per iteration.

\input{tab/priorbeta.tex}
\input{tab/priorcandidate.tex}
\input{tab/binarycontinous.tex}

\subsection{Discussion of Convergence}
\label{subsec:convergence}
As mentioned in Section~\ref{subsec:CA}, the synchronizing updating rule of the cellular automaton is designed to converge the evolved cells to a stable state after several generations. We compute the absolute difference of the $S_{\textup{Ref}}^{t}$ and $S_{\textup{Ref}}^{t-1}$ in Eq.~\ref{eq:sprior} at each generation, and plot the average absolute difference between $S_{\textup{Ref}}^{t}$ and $S_{\textup{Ref}}^{t-1}$ of all the superpixels on one image, with a combination of MB~\cite{zhang2015minimum}, BSCA~\cite{qin2015saliency}, RC~\cite{cheng2015global}, GBVS~\cite{harel2006graph}, COV~\cite{erdem2013visual} and FT~\cite{achanta2009frequency} on ECSSD dataset. The result is illustrated on Figure~\ref{fig:fm}~(c). As is shown, the designed updating rule for cellular automaton can make the $S_{\textup{Ref}}^{t}$ rapidly converge within five iterations.

\subsection{Running Time}
We implement our method in MATLAB R2014b using a Windows desktop with an i5-3570 CPU at 3.40GHz. The running time of AMS-B on ECSSD dataset ranges from 1.28s (2-model-combination) to 1.32s (8-model-combination) per image, while AML-B ranges from 1.38s (2-model-combination) to 2.06s (8-model-combination) in average, without code optimization. The AML and AMS show comparable performances, but AML takes longer time in running the EM algorithm.

%% file: tab/ecssddataset_adpt.tex
\begin{table*}
\centering
\footnotesize
\begin{tabular}{|c|r|c|c|c|c|c|c|c|c|c|c|c|}
\hline
\multicolumn{2}{|c|}{Combination}&Top&AVE &BN & MCA& M-est &AMS-B& AMS-C & AMS-D & AML-B& AML-C & AML-D \\\cline{3-13}
\hline\hline

\multirow{14}{*}{\rotatebox{90}{\textbf{Superior Models Combinations}}}&\multirow{2}{*}{\ssmall DRFI, MB+}&\multirow{2}{*}{\small0.765}&\multirow{2}{*}{\small0.750}&\multirow{2}{*}{\small0.750}&\multirow{2}{*}{\small0.760}&\multirow{2}{*}{\small0.752}&\multirow{2}{*}{\small0.791}&\multirow{2}{*}{\small0.794}&\multirow{2}{*}{\small\textbf{0.832}}&\multirow{2}{*}{\small0.785}&\multirow{2}{*}{\small\underline{0.803}}&\cellcolor{gray!20}\\
&&&&&&&&&&&&\multirow{-2}{*}{\cellcolor{gray!20}\small\textbf{0.850}}\\\cline{2-2}

&\multirow{2}{*}{\ssmall DRFI, MB+, RB}&\multirow{2}{*}{\small0.765}&\multirow{2}{*}{\small0.752}&\multirow{2}{*}{\small0.752}&\multirow{2}{*}{\small0.765}&\multirow{2}{*}{\small0.754}&\multirow{2}{*}{\small0.772}&\multirow{2}{*}{\small0.780}&\multirow{2}{*}{\small\textbf{0.827}}&\multirow{2}{*}{\small0.772}&\multirow{2}{*}{\small\underline{0.788}}&\cellcolor{gray!20}\\
&&&&&&&&&&&&\multirow{-2}{*}{\cellcolor{gray!20}\small\textbf{0.851}}\\\cline{2-2}

&\multirow{2}{*}{\ssmall DRFI, MB+, RB, TLLT}&\multirow{2}{*}{\small0.765}&\multirow{2}{*}{\small0.762}&\multirow{2}{*}{\small0.762}&\multirow{2}{*}{\small0.765}&\multirow{2}{*}{\small0.764}&\multirow{2}{*}{\small0.773}&\multirow{2}{*}{\small\underline{0.784}}&\multirow{2}{*}{\small\textbf{0.821}}&\multirow{2}{*}{\small0.766}&\multirow{2}{*}{\small0.781}&\cellcolor{gray!20}\\
&&&&&&&&&&&&\multirow{-2}{*}{\cellcolor{gray!20}\small\textbf{0.830}}\\\cline{2-2}

&{\ssmall DRFI, MB+, RB, TLLT}&\multirow{2}{*}{\small0.765}&\multirow{2}{*}{\small0.753}&\multirow{2}{*}{\small0.753}&\multirow{2}{*}{\small0.758}&\multirow{2}{*}{\small0.755}&\multirow{2}{*}{\small0.770}&\multirow{2}{*}{\small0.775}&\multirow{2}{*}{\small\textbf{0.819}}&\multirow{2}{*}{\small0.768}&\multirow{2}{*}{\small\underline{0.780}}&\cellcolor{gray!20}\\
&\ssmall MB&&&&&&&&&&&\multirow{-2}{*}{\cellcolor{gray!20}\small\textbf{0.836}}\\\cline{2-2}

&{\ssmall DRFI, MB+, RB, TLLT}&\multirow{2}{*}{\small0.765}&\multirow{2}{*}{\small0.757}&\multirow{2}{*}{\small0.757}&\multirow{2}{*}{\small0.762}&\multirow{2}{*}{\small0.759}&\multirow{2}{*}{\small0.778}&\multirow{2}{*}{\small0.781}&\multirow{2}{*}{\small\textbf{0.816}}&\multirow{2}{*}{\small0.774}&\multirow{2}{*}{\small\underline{0.784}}&\cellcolor{gray!20}\\
&\ssmall MB, BSCA&&&&&&&&&&&\multirow{-2}{*}{\cellcolor{gray!20}\small\textbf{0.832}}\\\cline{2-2}

&{\ssmall DRFI, MB+, RB, TLLT}&\multirow{2}{*}{\small0.765}&\multirow{2}{*}{\small0.764}&\multirow{2}{*}{\small0.764}&\multirow{2}{*}{\small0.767}&\multirow{2}{*}{\small0.766}&\multirow{2}{*}{\small0.783}&\multirow{2}{*}{\small0.783}&\multirow{2}{*}{\small\textbf{0.821}}&\multirow{2}{*}{\small0.778}&\multirow{2}{*}{\small\underline{0.787}}&\cellcolor{gray!20}\\
&\ssmall MB, BSCA, RC&&&&&&&&&&&\multirow{-2}{*}{\cellcolor{gray!20}\small\textbf{0.837}}\\\cline{2-2}

&{\ssmall DRFI, MB+, RB, TLLT}&\multirow{2}{*}{\small0.765}&\multirow{2}{*}{\small0.765}&\multirow{2}{*}{\small0.765}&\multirow{2}{*}{\small0.767}&\multirow{2}{*}{\small0.768}&\multirow{2}{*}{\small0.783}&\multirow{2}{*}{\small0.785}&\multirow{2}{*}{\small\textbf{0.815}}&\multirow{2}{*}{\small0.781}&\multirow{2}{*}{\small\underline{0.788}}&\cellcolor{gray!20}\\
&\ssmall MB, BSCA, RC, MR&&&&&&&&&&&\multirow{-2}{*}{\cellcolor{gray!20}\small\textbf{0.828}}
\\\hline

\multirow{14}{*}{\rotatebox{90}{\textbf{Inferior Models Combinations}}}&\multirow{2}{*}{\ssmall SR, IS}&\multirow{2}{*}{\small0.416}&\multirow{2}{*}{\small0.420}&\multirow{2}{*}{\small0.420}&\multirow{2}{*}{\small0.442}&\multirow{2}{*}{\small0.420}&\multirow{2}{*}{\small0.556}&\multirow{2}{*}{\small0.613}&\multirow{2}{*}{\small\textbf{0.688}}&\multirow{2}{*}{\small0.576}&\multirow{2}{*}{\small\underline{0.637}}&\cellcolor{gray!20}\\
&&&&&&&&&&&&\multirow{-2}{*}{\cellcolor{gray!20}\small\textbf{0.723}}\\\cline{2-2}

&\multirow{2}{*}{\ssmall CA, SR, IS}&\multirow{2}{*}{\small0.483}&\multirow{2}{*}{\small0.456}&\multirow{2}{*}{\small0.456}&\multirow{2}{*}{\small0.472}&\multirow{2}{*}{\small0.456}&\multirow{2}{*}{\small0.554}&\multirow{2}{*}{\small0.609}&\multirow{2}{*}{\small\textbf{0.690}}&\multirow{2}{*}{\small0.579}&\multirow{2}{*}{\small\underline{0.647}}&\cellcolor{gray!20}\\
&&&&&&&&&&&&\multirow{-2}{*}{\cellcolor{gray!20}\small\textbf{0.738}}\\\cline{2-2}

&\multirow{2}{*}{\ssmall FT, CA, SR, IS}&\multirow{2}{*}{\small0.483}&\multirow{2}{*}{\small0.465}&\multirow{2}{*}{\small0.465}&\multirow{2}{*}{\small0.491}&\multirow{2}{*}{\small0.465}&\multirow{2}{*}{\small0.550}&\multirow{2}{*}{\small0.612}&\multirow{2}{*}{\small\textbf{0.677}}&\multirow{2}{*}{\small0.567}&\multirow{2}{*}{\small\underline{0.635}}&\cellcolor{gray!20}\\
&&&&&&&&&&&&\multirow{-2}{*}{\cellcolor{gray!20}\small\textbf{0.715}}\\\cline{2-2}

&{\ssmall FT, CA, SR, IS}&\multirow{2}{*}{\small0.520}&\multirow{2}{*}{\small0.492}&\multirow{2}{*}{\small0.492}&\multirow{2}{*}{\small0.513}&\multirow{2}{*}{\small0.492}&\multirow{2}{*}{\small0.585}&\multirow{2}{*}{\small0.636}&\multirow{2}{*}{\small\textbf{0.711}}&\multirow{2}{*}{\small0.602}&\multirow{2}{*}{\small\underline{0.658}}&\cellcolor{gray!20}\\
&\ssmall IT&&&&&&&&&&&\multirow{-2}{*}{\cellcolor{gray!20}\small\textbf{0.744}}\\\cline{2-2}

&{\ssmall FT, CA, SR, IS}&\multirow{2}{*}{\small0.622}&\multirow{2}{*}{\small0.537}&\multirow{2}{*}{\small0.537}&\multirow{2}{*}{\small0.554}&\multirow{2}{*}{\small0.537}&\multirow{2}{*}{\small0.632}&\multirow{2}{*}{\small0.670}&\multirow{2}{*}{\small\textbf{0.734}}&\multirow{2}{*}{\small0.627}&\multirow{2}{*}{\small\underline{0.674}}&\cellcolor{gray!20}\\
&\ssmall LR, IT&&&&&&&&&&&\multirow{-2}{*}{\cellcolor{gray!20}\small\textbf{0.750}}\\\cline{2-2}

&{\ssmall FT, CA, SR, IS}&\multirow{2}{*}{\small0.622}&\multirow{2}{*}{\small0.562}&\multirow{2}{*}{\small0.562}&\multirow{2}{*}{\small0.574}&\multirow{2}{*}{\small0.562}&\multirow{2}{*}{\small0.669}&\multirow{2}{*}{\small0.692}&\multirow{2}{*}{\small\textbf{0.756}}&\multirow{2}{*}{\small0.662}&\multirow{2}{*}{\small\underline{0.702}}&\cellcolor{gray!20}\\
&\ssmall GBVS, LR, IT&&&&&&&&&&&\multirow{-2}{*}{\cellcolor{gray!20}\small\textbf{0.783}}\\\cline{2-2}

&{\ssmall FT, CA, SR, IS}&\multirow{2}{*}{\small0.622}&\multirow{2}{*}{\small0.583}&\multirow{2}{*}{\small0.583}&\multirow{2}{*}{\small0.595}&\multirow{2}{*}{\small0.583}&\multirow{2}{*}{\small0.685}&\multirow{2}{*}{\small0.703}&\multirow{2}{*}{\small\textbf{0.755}}&\multirow{2}{*}{\small0.675}&\multirow{2}{*}{\small\underline{0.707}}&\cellcolor{gray!20}\\
&\ssmall PCAS, GBVS, LR, IT&&&&&&&&&&&\multirow{-2}{*}{\cellcolor{gray!20}\small\textbf{0.778}}
\\\hline

\multirow{14}{*}{\rotatebox{90}{\textbf{Random Combinations}}}&\multirow{2}{*}{\ssmall DRFI, GP}&\multirow{2}{*}{\small0.765}&\multirow{2}{*}{\small0.768}&\multirow{2}{*}{\small0.768}&\multirow{2}{*}{\small0.776}&\multirow{2}{*}{\small0.770}&\multirow{2}{*}{\small0.793}&\multirow{2}{*}{\small0.794}&\multirow{2}{*}{\small\textbf{0.826}}&\multirow{2}{*}{\small0.789}&\multirow{2}{*}{\small\underline{0.803}}&\cellcolor{gray!20}\\
&&&&&&&&&&&&\multirow{-2}{*}{\cellcolor{gray!20}\small\textbf{0.846}}\\\cline{2-2}

&\multirow{2}{*}{\ssmall HS, IT, IS}&\multirow{2}{*}{\small0.700}&\multirow{2}{*}{\small0.636}&\multirow{2}{*}{\small0.636}&\multirow{2}{*}{\small0.592}&\multirow{2}{*}{\small0.633}&\multirow{2}{*}{\small0.708}&\multirow{2}{*}{\small0.719}&\multirow{2}{*}{\small\textbf{0.776}}&\multirow{2}{*}{\small0.705}&\multirow{2}{*}{\small\underline{0.745}}&\cellcolor{gray!20}\\
&&&&&&&&&&&&\multirow{-2}{*}{\cellcolor{gray!20}\small\textbf{0.835}}\\\cline{2-2}

&\multirow{2}{*}{\ssmall HS, GC, COV, CA}&\multirow{2}{*}{\small0.700}&\multirow{2}{*}{\small0.647}&\multirow{2}{*}{\small0.647}&\multirow{2}{*}{\small0.585}&\multirow{2}{*}{\small0.647}&\multirow{2}{*}{\small0.701}&\multirow{2}{*}{\small\underline{0.719}}&\multirow{2}{*}{\small\textbf{0.765}}&\multirow{2}{*}{\small0.709}&\multirow{2}{*}{\small0.715}&\cellcolor{gray!20}\\
&&&&&&&&&&&&\multirow{-2}{*}{\cellcolor{gray!20}\small\textbf{0.779}}\\\cline{2-2}

&{\ssmall HS, GC, COV, CA}&\multirow{2}{*}{\small0.735}&\multirow{2}{*}{\small0.727}&\multirow{2}{*}{\small0.728}&\multirow{2}{*}{\small0.723}&\multirow{2}{*}{\small0.729}&\multirow{2}{*}{\small0.736}&\multirow{2}{*}{\small0.755}&\multirow{2}{*}{\small\textbf{0.796}}&\multirow{2}{*}{\small0.735}&\multirow{2}{*}{\small\underline{0.772}}&\cellcolor{gray!20}\\
&\ssmall MR&&&&&&&&&&&\multirow{-2}{*}{\cellcolor{gray!20}\small\textbf{0.828}}\\\cline{2-2}

&{\ssmall MB, BSCA, RC, GBVS}&\multirow{2}{*}{\small0.736}&\multirow{2}{*}{\small0.733}&\multirow{2}{*}{\small0.734}&\multirow{2}{*}{\small0.743}&\multirow{2}{*}{\small0.734}&\multirow{2}{*}{\small0.744}&\multirow{2}{*}{\small0.755}&\multirow{2}{*}{\small\textbf{0.790}}&\multirow{2}{*}{\small0.747}&\multirow{2}{*}{\small\underline{0.770}}&\cellcolor{gray!20}\\
&\ssmall COV, FT&&&&&&&&&&&\multirow{-2}{*}{\cellcolor{gray!20}\small\textbf{0.825}}\\\cline{2-2}

&{\ssmall MB+, GP, BSCA, RB}&\multirow{2}{*}{\small0.736}&\multirow{2}{*}{\small0.736}&\multirow{2}{*}{\small0.736}&\multirow{2}{*}{\small0.726}&\multirow{2}{*}{\small0.736}&\multirow{2}{*}{\small0.751}&\multirow{2}{*}{\small0.757}&\multirow{2}{*}{\small\textbf{0.794}}&\multirow{2}{*}{\small0.758}&\multirow{2}{*}{\small\underline{0.764}}&\cellcolor{gray!20}\\
&\ssmall GBVS, IT, IS&&&&&&&&&&&\multirow{-2}{*}{\cellcolor{gray!20}\small\textbf{0.826}}\\\cline{2-2}

&{\ssmall MB, BSCA, TLLT, GC}&\multirow{2}{*}{\small0.736}&\multirow{2}{*}{\small0.742}&\multirow{2}{*}{\small0.743}&\multirow{2}{*}{\small0.724}&\multirow{2}{*}{\small0.743}&\multirow{2}{*}{\small0.761}&\multirow{2}{*}{\small0.768}&\multirow{2}{*}{\small\textbf{0.801}}&\multirow{2}{*}{\small0.755}&\multirow{2}{*}{\small\underline{0.770}}&\cellcolor{gray!20}\\
&\ssmall PCAS, GBVS, HS, CA&&&&&&&&&&&\multirow{-2}{*}{\cellcolor{gray!20}\small\textbf{0.822}}
\\\hline

\multirow{14}{*}{\rotatebox{90}{\textbf{Deep Models Combinations}}}&\multirow{2}{*}{\ssmall \underline{DCL}, \underline{DSS}}&\multirow{2}{*}{\small0.888}&\multirow{2}{*}{\small0.875}&\multirow{2}{*}{\small0.875}&\multirow{2}{*}{\small0.590}&\multirow{2}{*}{\small0.876}&\multirow{2}{*}{\small0.890}&\multirow{2}{*}{\small\underline{0.894}}&\multirow{2}{*}{\small\textbf{0.894}}&\multirow{2}{*}{\small0.886}&\multirow{2}{*}{\small0.893}&\cellcolor{gray!20}\\
&&&&&&&&&&&&\multirow{-2}{*}{\cellcolor{gray!20}\small\textbf{0.903}}\\\cline{2-2}

&\multirow{2}{*}{\ssmall \underline{DCL}, \underline{DSS}, \underline{RFCN}}&\multirow{2}{*}{\small0.888}&\multirow{2}{*}{\small0.837}&\multirow{2}{*}{\small0.838}&\multirow{2}{*}{\small0.814}&\multirow{2}{*}{\small0.844}&\multirow{2}{*}{\small0.894}&\multirow{2}{*}{\small\underline{0.895}}&\multirow{2}{*}{\small\textbf{0.898}}&\multirow{2}{*}{\small0.874}&\multirow{2}{*}{\small0.891}&\cellcolor{gray!20}\\
&&&&&&&&&&&&\multirow{-2}{*}{\cellcolor{gray!20}\small\textbf{0.902}}\\\cline{2-2}

&\multirow{2}{*}{\ssmall \underline{DCL}, \underline{DSS}, \underline{RFCN}, \underline{MDF}}&\multirow{2}{*}{\small0.888}&\multirow{2}{*}{\small0.845}&\multirow{2}{*}{\small0.845}&\multirow{2}{*}{\small0.816}&\multirow{2}{*}{\small0.852}&\multirow{2}{*}{\small0.887}&\multirow{2}{*}{\small\underline{0.892}}&\multirow{2}{*}{\small\textbf{0.895}}&\multirow{2}{*}{\small0.874}&\multirow{2}{*}{\small0.886}&\cellcolor{gray!20}\\
&&&&&&&&&&&&\multirow{-2}{*}{\cellcolor{gray!20}\small\textbf{0.898}}\\\cline{2-2}

&{\ssmall \underline{DCL}, \underline{DSS}, \underline{RFCN}, \underline{MDF}}&\multirow{2}{*}{\small0.888}&\multirow{2}{*}{\small0.861}&\multirow{2}{*}{\small0.861}&\multirow{2}{*}{\small0.821}&\multirow{2}{*}{\small0.867}&\multirow{2}{*}{\small0.896}&\multirow{2}{*}{\small\textbf{0.903}}&\multirow{2}{*}{\small\underline{0.901}}&\multirow{2}{*}{\small0.892}&\multirow{2}{*}{\small0.897}&\cellcolor{gray!20}\\
&\ssmall \underline{DHSNet}&&&&&&&&&&&\multirow{-2}{*}{\cellcolor{gray!20}\small\textbf{0.905}}\\\cline{2-2}

&{\ssmall \underline{DCL}, \underline{DSS}, \underline{RFCN}, \underline{MDF}}&\multirow{2}{*}{\small0.888}&\multirow{2}{*}{\small0.842}&\multirow{2}{*}{\small0.842}&\multirow{2}{*}{\small0.799}&\multirow{2}{*}{\small0.851}&\multirow{2}{*}{\small0.890}&\multirow{2}{*}{\small\underline{0.891}}&\multirow{2}{*}{\small\textbf{0.896}}&\multirow{2}{*}{\small0.877}&\multirow{2}{*}{\small0.886}&\cellcolor{gray!20}\\
&\ssmall MB+&&&&&&&&&&&\multirow{-2}{*}{\cellcolor{gray!20}\small\textbf{0.899}}\\\cline{2-2}

&{\ssmall \underline{DCL}, \underline{DSS}, \underline{RFCN}, \underline{MDF}}&\multirow{2}{*}{\small 0.888}&\multirow{2}{*}{\small0.838}&\multirow{2}{*}{\small0.838}&\multirow{2}{*}{\small0.793}&\multirow{2}{*}{\small0.846}&\multirow{2}{*}{\small0.884}&\multirow{2}{*}{\small \underline{0.884}}&\cellcolor{gray!20}&\multirow{2}{*}{\small0.872}&\multirow{2}{*}{\small0.881}&\multirow{2}{*}{\small\textbf{0.890}}\\
&\ssmall MB+,GP&&&&&&&&\multirow{-2}{*}{\cellcolor{gray!20}\small\textbf{0.891}}&&&\\\cline{2-2}

&{\ssmall \underline{DCL}, \underline{DSS}, \underline{RFCN}, \underline{MDF}}&\multirow{2}{*}{\small0.888}&\multirow{2}{*}{\small0.822}&\multirow{2}{*}{\small0.822}&\multirow{2}{*}{\small0.785}&\multirow{2}{*}{\small0.828}&\multirow{2}{*}{\small\underline{0.876}}&\multirow{2}{*}{\small0.872}& \cellcolor{gray!20}&\multirow{2}{*}{\small0.864}&\multirow{2}{*}{\small0.865}&\multirow{2}{*}{\small\textbf{0.884}}\\
&\ssmall MB+,GP,MB&&&&&&&&\multirow{-2}{*}{\cellcolor{gray!20}\small\textbf{0.889}}&&&\\\cline{2-2}

&{\ssmall \underline{DCL}, \underline{DSS}, \underline{RFCN}, \underline{MDF}}&\multirow{2}{*}{\small 0.888}&\multirow{2}{*}{\small0.840}&\multirow{2}{*}{\small0.841}&\multirow{2}{*}{\small0.789}&\multirow{2}{*}{\small0.845}&\multirow{2}{*}{\small0.874}&\multirow{2}{*}{\small \underline{0.877}}&\cellcolor{gray!20}&\multirow{2}{*}{\small0.864}&\multirow{2}{*}{\small0.872}&\multirow{2}{*}{\small\textbf{0.886}}\\
&\ssmall MB+,BSCA,DRFI,TLLT&&&&&&&&\multirow{-2}{*}{\cellcolor{gray!20}\small\textbf{0.889}}&&&
\\\hline

\end{tabular}\\[1ex]
      \caption{Mean F-measure of the average saliency maps (AVE), the resulted BN, M-estimator (M-est), and MCA saliency maps and the resulted AMS and AML saliency maps. The subscripts ``B'', ``C'' and ``D'' represent the boundary-based reference map, contour-based reference map and deep-network-based reference map respectively. The first column shows the combination strategy, and for every combination the highest F-measure of the candidate saliency models are displayed in the ``Top'' column. The \colorbox{gray!20}{\textbf{best result}} for each combination is in bold with dark background color, the \textbf{second best} is in bold, and the \underline{third best} is underlined. The candidate models include BSCA~\cite{qin2015saliency}, CA~\cite{goferman2012context}, CEOS~\cite{mairon2014closer}, COV~\cite{erdem2013visual}, DRFI~\cite{jiang2013salient}, FT~\cite{achanta2009frequency}, GBVS~\cite{harel2006graph}, GC~\cite{cheng2015global}, GP~\cite{jiang2015generic}, HS~\cite{yan2013hierarchical}, IS~\cite{hou2012image}, IT~\cite{itti1998model}, LR~\cite{shen2012unified}, MB~\cite{zhang2015minimum}, MB+~\cite{zhang2015minimum}, MR~\cite{yang2013saliency}, PCAS~\cite{margolin2013makes}, RB~\cite{wei2012geodesic}, RC~\cite{cheng2015global}, SR~\cite{hou2007saliency}, TLLT~\cite{gong2015saliency}, UFO~\cite{jiang2013salient}, DSS~\cite{hou2017saliency}, DCL~\cite{li2016deep}, RFCN~\cite{dai2016r}, MDF~\cite{li2015visual}, and DHSNet~\cite{liu2016dhsnet}, of which deep models are underlined.}
    \label{tbl:ecssd}
\end{table*}

%% file: tab/5dataset.tex
\begin{table*}
\small
\begin{center}
    \begin{tabular}{|c|c|c|c|c|c|c|c|c|c|c|c|}
    \hline
    Dataset&Top&AVE &BN & M-est& MCA & AMS-B& AMS-C & AMS-D & AML-B& AML-C & AML-D\\\hline

ECSSD&    0.736&0.736&0.733&0.734&0.743&0.744&0.755&\textbf{0.790}&0.747&\underline{0.770}&\cellcolor{gray!20}\textbf{0.825}\\\cline{1-1}

ASD	&     0.885&0.872&0.867&0.868&0.886&0.898&0.906&\textbf{0.917}&0.896&\underline{0.912}&\cellcolor{gray!20}\textbf{0.928}\\\cline{1-1}

ImgSal&   0.515&0.497&0.494&0.495&0.528&0.571&0.590&\textbf{0.636}&0.600&\underline{0.626}&\cellcolor{gray!20}\textbf{0.690}\\\cline{1-1}

DUT-OMRON&0.546&0.560&0.556&0.556&0.571&0.602&0.616&\textbf{0.690}&0.621&\underline{0.637}&\cellcolor{gray!20}\textbf{0.749}\\\hline

    \end{tabular}
      \caption{Mean F-measure of the top saliency model, average saliency maps, BN, M-estimator (M-est), MCA model, and AM model with a combination of MB~\cite{zhang2015minimum}, BSCA~\cite{qin2015saliency}, RC~\cite{cheng2015global}, GBVS~\cite{harel2006graph}, COV~\cite{erdem2013visual} and FT~\cite{achanta2009frequency} models on four datasets including ECSSD~\cite{yan2013hierarchical}, ASD~\cite{achanta2009frequency}, ImgSal~\cite{li2013visual} and DUT-OMRON~\cite{yang2013saliency}. The highest F-measure of the candidate saliency models are displayed in the ``Top'' column. The \colorbox{gray!20}{\textbf{best result}} for each combination is in bold with dark background color, the \textbf{second best} is in bold, and the \underline{third best} is underlined.}
    \label{tbl:5dataset}
\end{center}
\end{table*}

%% file: tab/priorbeta.tex
\begin{table}
\begin{center}
   \small
    \begin{tabular}{r|c|c|c|c|c|c}
    \hline
    Ref-B       &   $\times$&   $\times$&   $\times$&     $\surd$&   $\surd$&  $\surd$ \\\hline
    Expertise   &   $\times$&       L&       S&    $\times$&      L&     S \\\hline
    F-Measure   &       0.757&       0.767&       0.762&        0.769&      0.777&     0.784 \\\hline
    \end{tabular}
      \caption{Mean F-measure of saliency integration framework incorporating various combination of the proposed components by AM model. The letter ``L'' refers to the latent-variable-based expertise and ``S'' means the statistics-based expertise. ``Ref-B'' means that we test the boundary-based reference map. ``$\surd$'' and ``$\times$'' indicate whether the component is incorporated into CA or not.}
    \label{tbl:priorexp}
\end{center}
\end{table} 

%% file: tab/priorcandidate.tex
\begin{table}
\begin{center}
   \small
    \begin{tabular}{r|c|c|c|c}
    \hline
    External Knowledge&    FT&      Bound&      CCM&   DHSNet \\\hline
    Reference    &    0.766&    0.789&   0.777&   0.837 \\
    Candidate&    0.744&    0.750&   0.757&   0.774 \\\hline
    \end{tabular}
      \caption{Mean F-measure of saliency integration results of the AML model by involving different external knowledge. The first row (``Reference'') and the second row (``Candidate'') refer to the F-measure by incorporating the external knowledge as the reference map and as one of the candidate maps to be aggregated, respectively.}
    \label{tbl:priorcandidate}
\end{center}
\end{table} 

%% file: tab/binarycontinous.tex
\begin{table}
\begin{center}
   \small
    \begin{tabular}{r|c|c|c|c|c|c}
    \cline{2-7}
    &\multicolumn{3}{c|}{AML}&\multicolumn{3}{c}{AMS}\\\cline{2-7}
    &    B&      C&      D&    B&      C&      D \\\hline
    Binary    & 0.777&0.787&0.833&0.781&0.788&0.822 \\
    Continuous &0.759&0.74&0.762&0.603&0.511&0.509 \\
    Both& 0.777&0.789&0.837&0.781&0.788&0.823 \\\hline
    \end{tabular}
      \caption{Mean F-measure of saliency integration by using AM model as framework with only continuous saliency maps, only binary maps, and both continuous and binary maps for integration. The candidate saliency models are DRFI, GP, LR, MB+, TLLT and UFO on ECSSD dataset.}
    \label{tbl:binary_continous}
\end{center}
\end{table} 

%% file: Conclusion.tex
\section{Conclusion}
This paper presents an arbitrator model (AM) as an efficient online saliency integration model to release the burden of model-training from offline models.
On one hand, the AM model introduces the reference map to overcome the misleading of inferior saliency models by exploring the consensus of the multiple saliency maps and the external knowledge. On the other hand, it rationally learns the expertise of saliency models without any knowledge of the ground truth labels in an online manner.
We evaluate the AM with a pool of twenty-seven models under various combinations. The experimental results show that it substantially improves the performance, regardless of the choices of candidate approaches.

We also hold discussions about the two proposed online methods in estimating the expertise of saliency models, namely the statistics-based expertise and the latent-variable-based expertise.
It can be easily observed that the statistics-based expertise is more accurate than the latent-variable-based one, if without the reference map. With the incorporation of the reference map, the two proposed expertise estimation methods perform similarly well. Nevertheless, the computational cost of the latent-variable-based method is higher than that of the statistics-based method, especially when the number of candidate models increases. Therefore,
the statistics-based expertise is more efficient than the latent-variable-based expertise when large numbers of saliency models are integrated.

The AM model proposes a new integration framework that incorporates the reference map and the candidate saliency models of varying expertise. Although the framework is derived from Bayesian inference, the computation of each component can be further investigated. Firstly, the design of each component can be explored for further gains in performance. This paper incorporates the consensus of multiple saliency models and the external knowledge to approximate the reference map. To further improve the quality of the reference map, the external knowledge can be introduced in more complex ways, \textit{i.e.}, using multiple external knowledge rather than one to increase the validity of the reference map. Also, the multiple saliency models could contribute to the reference map in other forms rather than the majority voting, such as adopting mean field approximation~\cite{krahenbuhl2011efficient} to estimate the reference map. Secondly, the expertise can also be evaluated in different forms. In this work, we suggest to use the latent-variable-based method if the quality of the reference map is poor and the statistics-based method if the reference map is powerful. However, the latent-variable-based approach and the statistics-based approach can also be combined to stabilize the accuracy of expertise estimation if the quality of the reference map is uncertain and the computational cost is out of consideration.

Currently, at the updating stage of the cellular automaton in AM model, the state of a cell is only affected by the superpixels at the same location of all the saliency maps. In future, we will explore the influences of the adjacent superpixels of a cell. Also, the reference map and the expertise estimation of the AM framework can also be applied to co-saliency detection~\cite{zhang2017co,yao2017revisiting}. Moreover, we could further apply the framework to other tasks such as anomaly detection~\cite{ponti2017diver}.